\documentclass[10pt,twocolumn,letterpaper]{article}

\usepackage[pagenumbers]{cvpr} 
\usepackage{algorithm}
\usepackage{algpseudocode}
\usepackage{amsmath}
\usepackage{bbding}
\usepackage{booktabs}
\usepackage{multirow}
\usepackage{listings}
\usepackage{colortbl}
\usepackage{xcolor}
\usepackage{pifont}
\usepackage{graphicx}
\definecolor{cvprblue}{rgb}{0.21,0.49,0.74}
\usepackage[pagebackref,breaklinks,colorlinks,allcolors=cvprblue]{hyperref}

\def\paperID{12149} 
\def\confName{CVPR}
\def\confYear{2025}

\usepackage{CJKutf8}

\title{
    \begin{minipage}[c]{0.15\textwidth} 
        \hspace{1cm}
        \includegraphics[width=1.4cm]{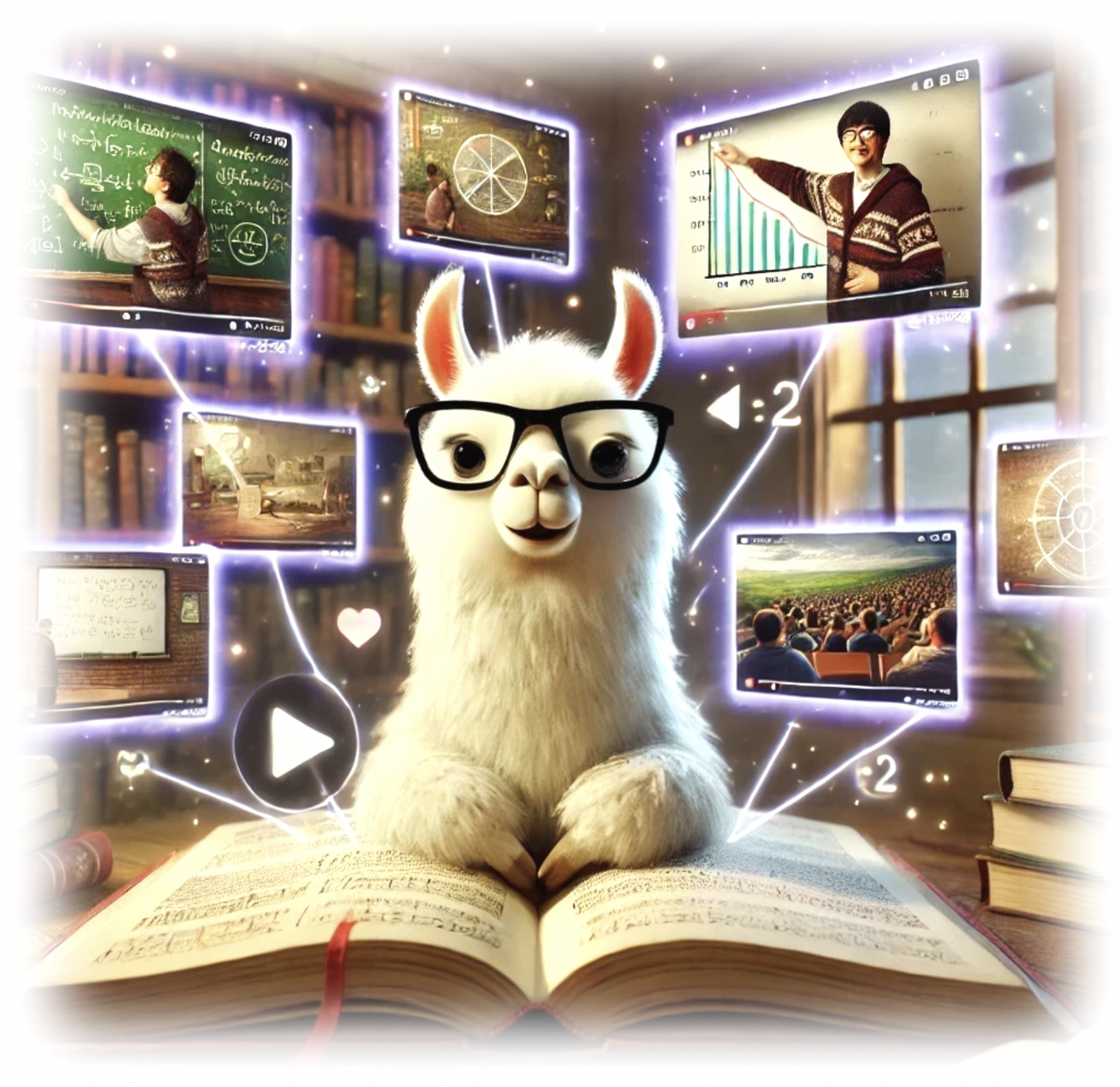} 
    \end{minipage}%
    \begin{minipage}[c]{0.8\textwidth} 
        \raggedright 
        2.5 Years in Class: \\A Multimodal Textbook for Vision-Language Pretraining
    \end{minipage}
}

\author{Wenqi Zhang\textsuperscript{1 *}
\hspace{0.3in} Hang Zhang\textsuperscript{2}
\hspace{0.3in} Xin Li\textsuperscript{2, $\dagger$}
\hspace{0.3in} Jiashuo Sun\textsuperscript{2} \\
\hspace{0.3in} Yongliang Shen\textsuperscript{1} 
\hspace{0.3in} Weiming Lu\textsuperscript{1, $\dagger$}
\hspace{0.3in} Deli Zhao\textsuperscript{2}
\hspace{0.3in} Yueting Zhuang\textsuperscript{1}
\hspace{0.3in} Lidong Bing\textsuperscript{2}
\vspace{10pt}
\\
\textsuperscript{1}{College of Computer Science and Technology, Zhejiang University} \\
\hspace{0.3in} \textsuperscript{2}{DAMO Academy, Alibaba Group} \\
zhangwenqi@zju.edu.cn  \\
\text{Project: \url{https://multimodal-interleaved-textbook.github.io/}} 
}

\begin{document}
\maketitle
\renewcommand{\thefootnote}{\fnsymbol{footnote}} 
\footnotetext[1]{This work was conducted when Wenqi Zhang was interning at Alibaba DAMO Academy.} 
\footnotetext[2]{Corresponding author.}  

\renewcommand{\thefootnote}{\arabic{footnote}}

\begin{abstract}


Compared to image-text pair data, interleaved corpora enable Vision-Language Models (VLMs) to understand the world more naturally like humans. However, such existing datasets are crawled from webpage, facing challenges like low knowledge density, loose image-text relations, and poor logical coherence between images. 
On the other hand, the internet hosts vast instructional videos (e.g., online geometry courses) that are widely used by humans to learn foundational subjects, yet these valuable resources remain underexplored in VLM training. In this paper, we introduce a high-quality \textbf{multimodal textbook} corpus with richer foundational knowledge for VLM pretraining. It collects over 2.5 years of instructional videos, totaling 22,000 class hours. We first use an LLM-proposed taxonomy to systematically gather instructional videos. Then we progressively extract and refine visual (keyframes), audio (ASR), and textual knowledge (OCR) from the videos, and organize as an image-text interleaved corpus based on temporal order.
Compared to its counterparts, our video-centric textbook offers more coherent context, richer knowledge, and better image-text alignment. Experiments demonstrate its superb pretraining performance, particularly in knowledge- and reasoning-intensive tasks like ScienceQA and MathVista. Moreover, VLMs pre-trained on our textbook exhibit outstanding interleaved context awareness, leveraging visual and textual cues in their few-shot context for task solving~\footnote{Our code are available at \url{https://github.com/DAMO-NLP-SG/multimodal_textbook}}.

\end{abstract}    
\section{Introduction} \label{introduction}
Vision-Language Models (VLMs) have demonstrated impressive development recently, delivering exceptional performance across a variety of visual tasks, including image captioning, dialogue, and visual question answering~\cite{blip2, awadalla2023openflamingo, achiam2023gpt, chen2023internvl, huang2023language, liu2024improved, team2024gemini, zhu2023minigpt, Qwen2VL,llava,sun2024generative,lu2024deepseek,laurenccon2024building,zhang2025embodied}. These advancements can be primarily attributed to the swift improvements of large language models (LLMs) and the community's ongoing creation of diverse, high-quality multimodal training corpora~\cite{schuhmann2022laion,jiang2024mantis,chen2024comm,awadalla2024mint, jia2021scaling,kakaobrain2022coyo}, collectively driving VLMs forward. A multimodal corpus typically consists of numerous image-text pairs to align images with textual descriptions. Pretraining on such paired datasets allows LLMs to be efficiently adapted into VLMs, with the ability to perceive and interpret visual information.

\begin{figure*}[t!]
  \centering
   \includegraphics[width=0.95\linewidth]{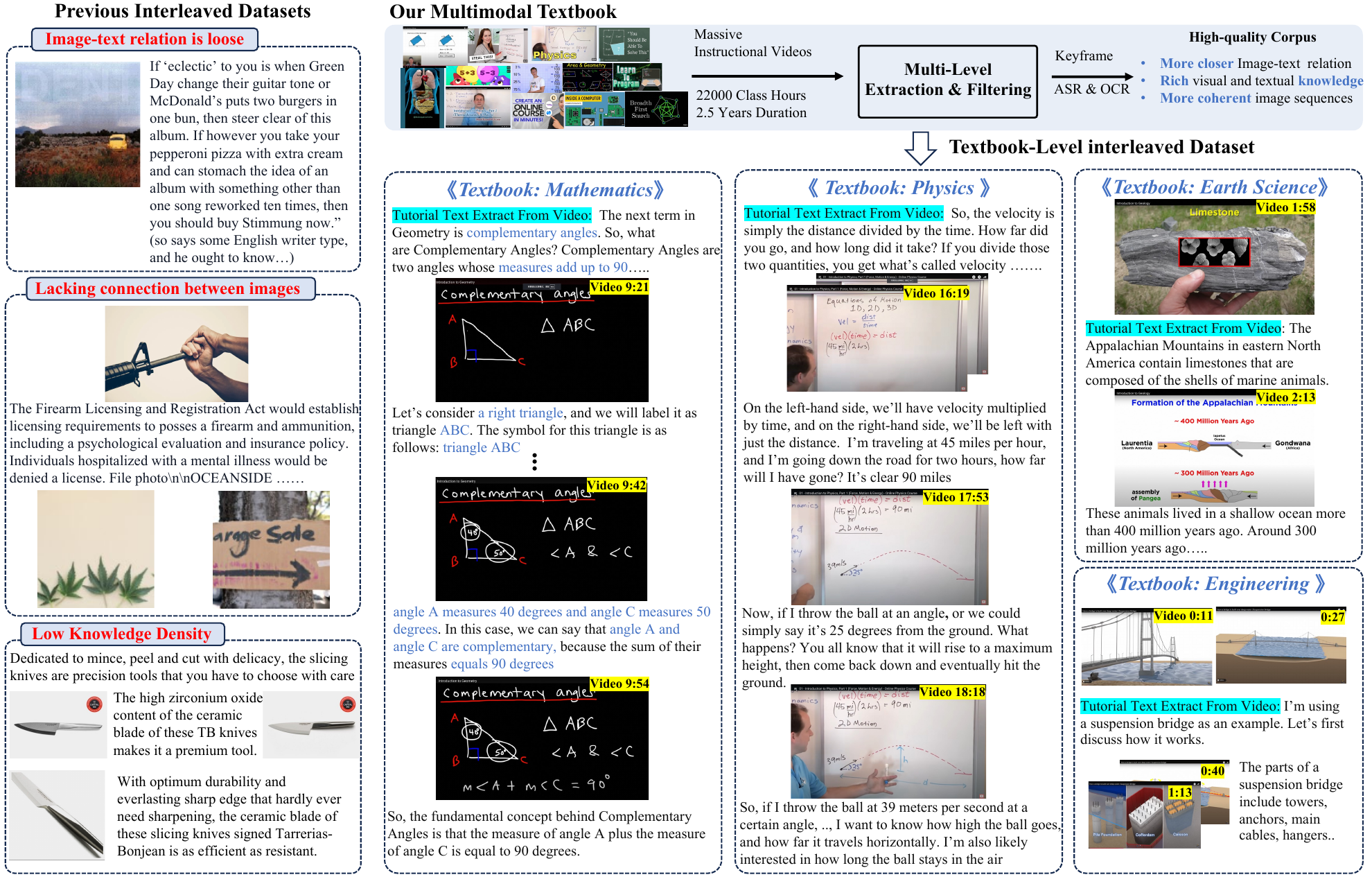}
   \caption{Previous interleaved datasets, e.g., MMC4 and OBELICS, suffer from limitations like weak text-image relations, low knowledge density, and incoherent image sequences. Our multimodal textbook, sourced from massive tutorial videos, employs coarse-to-fine knowledge extraction and multi-level filtering to create a high-quality, textbook-level dataset. It interleaves video keyframes with tutorial texts (extracted from ASR and OCR), enabling VLMs to acquire rich knowledge through tightly coupled text-image and more coherent logic.}
   \label{fig1}
\end{figure*}

Beyond image-text pair data, previous researchers have also introduced image-text interleaved corpus as a more natural and flexible multimodal corpus~\cite{alayrac2022flamingo,laurencon2023obelics,zhu2024multimodal,mckinzie2024mm1, li2024omnicorpus}. These corpora, consisting of sequences of text paragraphs interspersed with images, are typically crawled from webpage and document, such as Common Crawl. Pretraining on a combination of interleaved corpus and image-pair datasets enables VLMs to handle interwoven multi-modal inputs, while also unlocking advanced capabilities such as in-context learning~\cite{lin2023vila} and multi-image comparison~\cite{jiang2024mantis}.

Despite their benefits to multi-modal pre-training, existing interleaved datasets still suffer from the following issues (shown in~\cref{fig1}): (1) Loose text-image relation: The associations between images and text in a webpage are often loose and may even include irrelevant images, e.g., logos or advertisements. (2) Lack of logical coherence in image sequences: most webpages contain relatively few images, and more importantly, the logical relations between images are often vague, making it difficult to learn complex visual reasoning. (3) Low knowledge density: crawled webpages inevitably include content such as news, entertainment, and advertisement recommendations, with little involvement of fundamental knowledge. These issues may severely affect the learning effectiveness of interleaved corpora. Therefore, exploring how to extract high-quality, textbook-level interleaved datasets from vast internet data is quite necessary.

On the other hand, the internet contains a vast array of instructional videos~\cite{miech2019howto100m, sanabria2018how2, zellers2022merlot, hu2025video}, e.g., online mathematics courses on YouTube, where people often turn to acquire both foundational knowledge and specialized skills. Most videos contain frame-by-frame demonstrations along with detailed verbal explanations by the instructor, making them an ideal source of training data. However, these valuable resources have received limited attention for VLM training. Besides, Microsoft’s Phi-series models~\cite{gunasekar2023textbooks,li2023textbooks,javaheripi2023phi,abdin2024phi,abdin2024phi3} have also demonstrated that high-quality textbook-level datasets are critical for LLM training.

In this paper, we introduce a multimodal Textbook: a high-quality pre-training corpus that encompasses a wealth of foundational knowledge. Our textbook is constructed from 2.5 years of instructional videos, amounting to 22,000 class hours, covering six fundamental subjects, including mathematics, physics, and others. The whole corpus is presented in an image-text interleaved format, where the text and images are more closely aligned, and the logical relations between images are also more coherent.

To create our textbook, we develop an LLM-powered pipeline to systematically collect a vast array of instructional videos from the internet. To achieve automation, we prompt LLMs to construct a knowledge taxonomy covering six subjects and 3900 knowledge points. Then based on this, we gather relevant instructional videos. After that, we design a multi-level, coarse-to-fine knowledge extraction and data filtering pipeline for these collected videos. From a visual perspective, we extract keyframes and recognition text, symbols, and formulas (OCR). From an auditory perspective, we perform automatic speech recognition (ASR) on the instructor's verbal explanations and refine their quality. Finally, the keyframes and tutorial text are organized into an interleaved format, sequenced chronologically.

Our textbook is an openly accessible pre-training dataset with high-quality 6.5 million images interleaving with 0.75 billion texts. It drawn from 75,000 extensive instructional videos, totoaling over 22000 class hours, covering multiple core subjects such as mathematics, physics, chemistry. As demonstrated in~\cref{fig1}, our textbook (the first example) presents three keyframes interleaved with four tutorial texts to dynamically illustrate the geometric concept of complementary angles. These more coherent interleaved context and better-aligned image-text sequences enable VLMs to better grasp foundational knowledge during the pretraining.

Experiments show that VLMs pre-trained on our textbook achieve noticeable improvement on knowledge- and reasoning-intensive benchmarks, like MathVista, and ScienceQA. Besides, we also observe some intriguing findings: our textbook can significantly enhance the interleaved context awareness of VLMs, i.e., pretrained on our textbook, VLMs can more effectively attend to their few-shot context, leveraging visual or textual cues for question-solving. In contrast, the VLMs training on other datasets often overlooked their interleaved context.

\section{Related Works}
\label{related works}
\subsection{Vision Language Models}
With the development of large language models (LLMs)~\cite{gpt3,llama,yang2024qwen2}, VLMs have evolved from these task-specific, closed-set models~\cite{clip,li2020unicoder} to more flexible systems capable of handling open-world scenarios. Large VLMs adopt a general paradigm of mapping pretrained visual encoder outputs to the embedding space of LLMs, enabling cross-modal understanding~\cite{blip2,llava}. By leveraging large-scale caption datasets~\cite{laion400m,datacomp} and meticulously crafted instruction-following data~\cite{llava,cheng2024videollama}, these models exhibit remarkable capabilities. Building on this foundation, researchers have further boosted VLM performance by diversifying instruction data~\cite{yang2024qwen2,ye2024mplug}, refining data quality~\cite{gu2024infinity,lin2023vila}, and increasing image resolution~\cite{chen2023internvl,yao2024minicpm}. These improvements have led to breakthroughs across OCR, VQA, and visual grounding tasks, with VLMs now achieving impressive results on benchmarks that demand precise, context-aware understanding~\cite{chen2023internvl,lin2023vila,liu2024improved,yue2024mmmu}.

\subsection{Multi-modal Pretraining Data}

Recent developments in Vision-Language Models have typically involved a two-stage process: pretraining followed by a high-quality instruction-following phase~\cite{Qwen2VL,Qwen-VL,chen2024far,chen2023internvl,yao2024minicpm,liu2023improvedllava,liu2023llava,damonlpsg2023videollama}. 
Most VLMs utilize paired image-caption datasets~\cite{datacomp,schuhmann2022laion,laion400m} for pretraining which facilitate a quick alignment between image and text spaces~\cite{chen2023internvl,yao2024minicpm,liu2023improvedllava}. 
However, image-caption datasets lack the naturalness and authenticity found in more comprehensive text corpora used for LLMs, as they are often limited in diversity and complexity~\cite{lin2023vila}. This limitation reduces VLMs' capacity for in-context learning and chain-of-thought (CoT) reasoning. Recognizing this gap, some researchers have introduced webpage-centric interleaved datasets, like MMC4~\cite{zhu2024multimodal} and OBELICS~\cite{laurencon2023obelics}, sourced from webpages and documents~\cite{alayrac2022flamingo,awadalla2023openflamingo}. These interleaved datasets can enhance in-context learning capabilities in VLMs~\cite{lin2023vila,wang2024pin}. However, these datasets still face issues such as low image-text relevance, poor sequence logic, and sparse knowledge density. Our work proposes a multimodal ``textbook" corpus curated from instructional videos, intending to enhance multimodal pretraining and expand the model’s ability to handle interleaved visual and textual inputs.

\begin{figure*}[t!]
  \centering
   \includegraphics[width=0.9\linewidth]{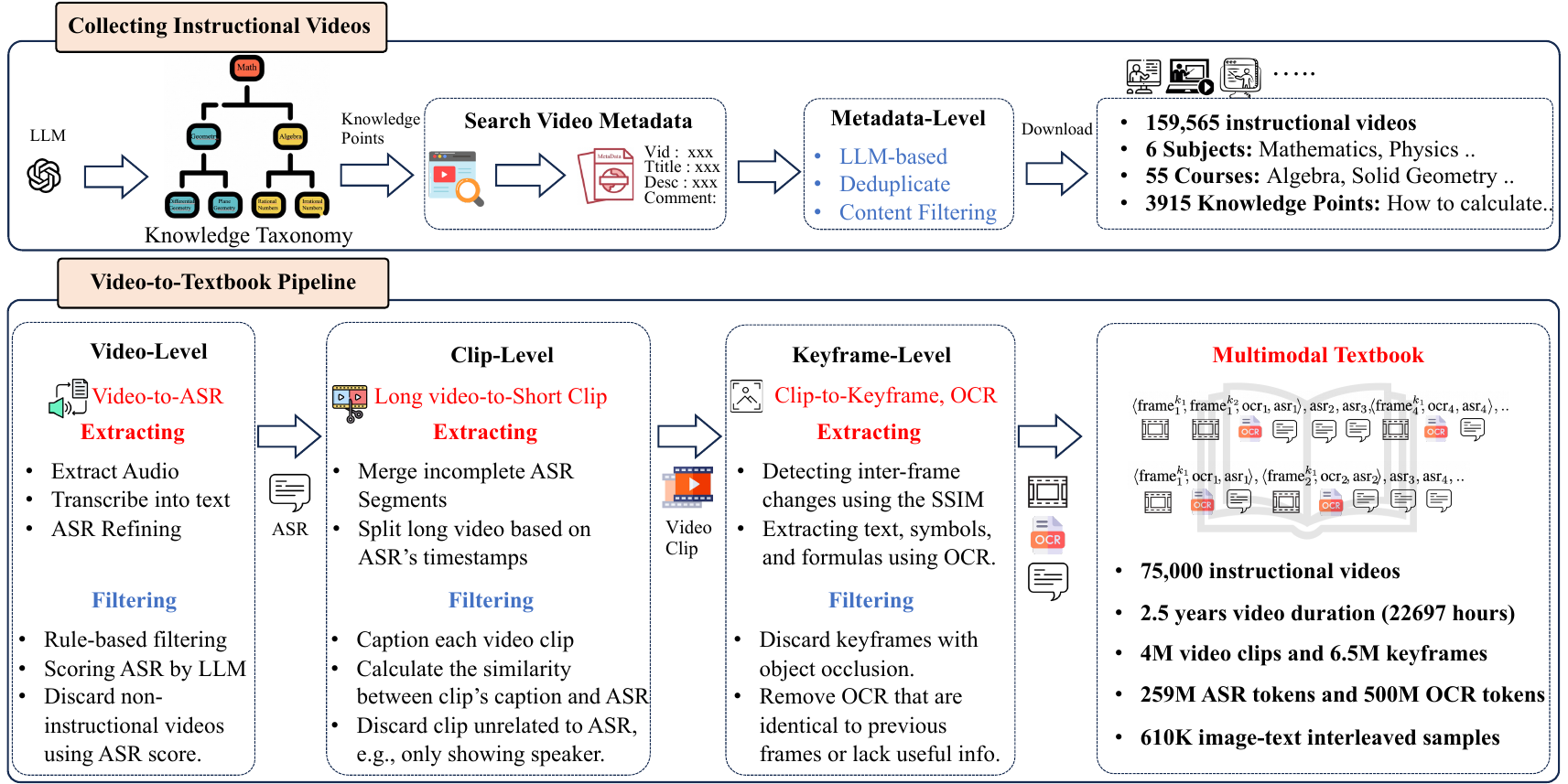}
   \caption{An illustration of constructing a multimodal textbook from instructional videos.  We first instruct LLMs to construct a knowledge taxonomy, then retrieve and filter videos at metadata level, collecting 159K instructional videos. Then a video-to-textbook pipeline is designed for multi-level knowledge extraction. \ding{172} We filter out non-instructional videos using ASR transcripts, retaining 75K high-quality videos. \ding{173} We use ASR's timestamp to segment long videos into short clips, discarding those with misaligned visuals and ASR. \ding{174} We detect keyframes from each clip and extract text and symbols by OCR. Our pipeline produces 6.5M keyframes, 259M ASR, and 500M OCR tokens and organizes them into an image-text interleaved textbook.}
   \label{fig2}
\end{figure*}


\section{Curation of Multimodal Textbook} \label{pipeline}
Our goal is to construct a textbook-level interleaved corpus that delivers high-quality, specialized knowledge for pre-training VLMs in a more natural and efficient manner.  To achieve this, we choose online instructional videos as the primary data source. Compared to common videos, such as entertainment, sports, or TV-show, instructional videos exhibit greater textual-visual consistency and sequential frame coherence, making them ideal for creating a “multimodal textbook”. While these videos are generally reliable, they still contain significant noise and redundancy, such as unrelated segments (e.g., advertisements), mismatches between visual content and text (e.g., almost ``static" scene predominantly featuring a single lecturer), or redundant scenes. To address this, we employ a multi-level pipeline (video-level, clip-level, and keyframe-level) with a coarse-to-fine strategy. The curation process is outlined in~\cref{fig2}.

\subsection{Collecting Instructional Videos} \label{collect video}
\textbf{LLM-proposed Knowledge Taxonomy.} In this work, we propose a knowledge taxonomy with four hierarchical layers for the desired instructional videos, namely \emph{Subject $\rightarrow$ Course $\rightarrow$ Sub-course $\rightarrow$ Knowledge Point}. To guarantee a broad coverage of instructional videos, we instruct an LLM to span the proposed knowledge taxonomy so that multiple educational stages (from primary school to middle school) and diverse subjects (mathematics, physics, etc.) will be involved. Eventually, as shown in~\cref{Taxonomy}, we obtain a knowledge taxonomy comprising 6 subjects (mathematics, physics, chemistry, earth science, engineering, and computer science), 55 courses (Algebra, Solid Geometry,..), and 3915 knowledge points. For example in the mathematics: \emph{Mathematics} $\rightarrow$ \emph{Elementary Mathematics} $\rightarrow$ \emph{Rational and Irrational Numbers} $\rightarrow$ \emph{the definition of Irrational Numbers}.

\textbf{Taxonomy-based Video Collection and Filtering.} Each knowledge point in the taxonomy is then used as a keyword to retrieve relevant instructional videos via YouTube's search API\footnote{\url{https://www.youtube.com/}}. We retain the top 50 videos for each knowledge point. Then, we perform deduplication based on video IDs and filter the low-quality videos using their metadata: we prompt LLMs to review each video's metadata—including the title, description, and comments—to exclude irrelevant, pornographic, or illegal content. Lastly, we collect a total of 159,565 videos from YouTube.

\subsection{Video-to-Textbook Pipeline} \label{pipeline}
For an instructional video, both the visual content (e.g., slide or animation) and the auditory content (e.g., instructor's narration) contain valuable knowledge. Therefore, we design a multi-level extraction pipeline to gather instructional keyframes and text from raw videos, interleaving them into a textbook.

\textbf{Video-Level Extraction: Video-to-ASR.} We employ \texttt{FFmpeg}\footnote{\url{https://www.ffmpeg.org/}} to extract the audio from each video (video-to-audio) and then transcribe it into text (audio-to-text, ASR) using \texttt{whisper-large-v3}\footnote{\url{https://huggingface.co/openai/whisper-large-v3}}. These transcriptions contain substantial knowledge and reasoning details, such as the instructor's explanations of on-screen content and step-by-step derivations of specific mathematical concepts. However, due to the nature of tutorial speech where the instructors prefer to use colloquial expressions to explain a concept, the perplexities (PPLs) of the raw ASR transcriptions are usually much higher than those of the texts from standard corpora (see~\cref{ablation study} for the concrete numbers). Therefore, we further introduce \texttt{Qwen2-72B-Instruct}~\citep{yang2024qwen2} to rewrite the raw ASR transcriptions, with the purpose of improving their fluency and coherence while not changing the original semantics

\textbf{Video-Level Filtering: Low-quality Videos based on ASR.} We first filter the videos using a set of predefined rules, including non-English videos, videos shorter than 10 seconds, and silent videos with very few ASR text tokens. Next, we assess the remaining videos by instructing an LLM to review their ASR transcriptions and filter out the non-instructional videos in terms of the following criteria: 
\begin{itemize}
    \item \textit{Relevance}: The ASR represents the tutorial content of the video. We assess the alignment between the ASR and the targeted knowledge point, filtering out irrelevant videos, e.g., advertisements or entertainment videos.
    \item \textit{Knowledge Density}: We evaluate the knowledge involved in ASR, as many videos contain meaningless filler phrases like "um," "the next up is this," or "then we get this." Such videos fail to provide valuable textual knowledge and are therefore discarded.
    \item \textit{Transcription Quality}: We examine the transcription quality by whisper, excluding repetitive or erroneous ASR text. This step occurs before ASR rewriting.
\end{itemize}

\begin{table*}[t!]
    \centering 
    \scriptsize
    \setlength\tabcolsep{6pt} 
    \begin{tabular}{l|lcc|ccc|cccccc|c}
        \toprule[1pt]
        \multirow{2}{*}{Dataset} & \multicolumn{3}{c}{\#Image} & \multicolumn{3}{c}{\#Text Token} & \multicolumn{6}{c}{$\text{In-sample Image SIM}^{L}$ $\uparrow$} & Source\\ 
         & \text{Min.} &\text{Max.} & \text{Avg.} & \text{Min.} &\text{Max.} & \text{Avg.} & \emph{L}=4& \emph{L}=5& \emph{L}=6 & \emph{L}=7& \emph{L}=8  & \emph{Avg.} & Common Crawl\\ \hline
        \multicolumn{8}{l}{\emph{Image-text Paired Dataset}}  \\ \hline
        COYO-700M  &1 &1 &1 &1 &811 & 16  & - &- &- &-&-&- &Common Crawl\\  
        LAION-5B &1 &1 &1 &6 &683  & 27  & - &- &- &-&-& -&Common Crawl  \\ \hline
        \multicolumn{8}{l}{\emph{Image-text Interleaved Dataset}}  \\ \hline 
        MMC4  &0 &117 &5.7 & 4 &16715 &417  & 0.363 & 0.348 & 0.310&0.298 &0.276 & 0.319  & Common Crawl\\
        MMC4-core-ff &0 &15 &4.1 &15 &16715 &329 &0.431& 0.406 & 0.404&	0.403	&0.396&	0.407 &Common Crawl\\
        OBELICS  &1 &30 &2.5 & 12 & 10717& 816 & 0.366 & 0.351&	0.339 &	0.337	&0.336	& 0.345  & Common Crawl\\    
        $\text{OmniCorpus}^{*}$   &1 &16 &3.9 &14 &6893 &574 & 0.358 & 0.329 & 0.310&0.305&0.301& 0.321& Multi-sources \\ \midrule
        \textbf{Ours}  & \textbf{2} & 45 & \textbf{10.7} & 11 & \textbf{34174} & \textbf{1297} & \textbf{0.687}	&\textbf{0.697}	&\textbf{0.698}	&\textbf{0.688}	&\textbf{0.662}	& \textbf{0.686} &Video Website\\ 
        \bottomrule[1pt]
    \end{tabular}    
    \caption{We compare our multimodal textbook with image-text paired datasets and webpage-centric interleaved datasets in terms of image and text distributions. In-sample Image $\text{SIM}^{L}$ measures the semantic and structural correlation between multiple images within an interleaved sample. $\text{OmniCorpus}^{*}$: Due to the extensive size of the dataset, we perform statistical analysis on a randomly sampled subset.}
    \label{Dataset compare}
\end{table*}

After LLM evaluation across these three dimensions, the retained 75,000 videos are generally of high quality, as verified by their ASR transcriptions.

\textbf{Clip-Level Extraction: Long Video-to-Short Clips.} To achieve temporal alignment between text and frames, we use the timestamps of each ASR transcription to segment the long video into multiple video clips. However, it is essential to consider that the original ASR transcriptions are often fragmented. First, we merge multiple incomplete ASR segments into a single, semantically coherent paragraph. Then, we use their timestamps to segment the video clips accordingly. Each clip lasts 10 to 20 seconds, accompanying an ASR text segment: $\left\langle \text{clip}_1, \text{asr}_1 \right\rangle, \left\langle \text{clip}_2, \text{asr}_2 \right\rangle, \dots, \left\langle \text{clip}_n, \text{asr}_n \right\rangle$

\textbf{Clip-Level Filtering: Video Clips without Visual Knowledge.} Previous filtering of long videos is based on ASR text. Next, we also assess each video clip from a visual perspective to determine if it contains sufficient visual knowledge. In most videos, it is inevitable to contain uninformative scenes, such as transitions, shots focused solely on the speaker, or cluttered backgrounds, which are not suitable for a multimodal textbook. A good scene should contain slides, blackboards, or demonstrative animations that introduce a knowledge concept or illustrate specific objects, rather than just the speaker alone. To this end, we employ a VideoLlama2~\citep{cheng2024videollama} to generate a detailed caption for each video clip. We then calculate the text similarity between the clip's caption and ASR transcription using the text embeddings model (gte-Qwen2-7B-instruct~\cite{li2023towards}), filtering out uninformative video clips.

Notably, even if an uninformative video clip is discarded, its ASR transcription may still contain valuable information. Thus, we retain these transcriptions in our textbook: $\left\langle \text{clip}_1, \text{asr}_1 \right\rangle,  \text{asr}_2, \text{asr}_3, \left\langle \text{clip}_4, \text{asr}_4 \right\rangle, \dots, \left\langle \text{clip}_n, \text{asr}_n \right\rangle$

\textbf{Keyframe-Level Extraction: Clip-to-Keyframes by Comparing Changes between Consecutive Two Frames.} Then we need to extract keyframes from each video clip, removing similar or even duplicate shots. A frame is identified as a keyframe if it exhibits significant visual change compared to the previous one. Therefore, we compute the similarity between consecutive frames and filter out those with minimal scene changes.

Considering efficiency and accuracy, we employ the Structural Similarity Index algorithm (SSIM)~\citep{wang2004image} to compare the similarity between consecutive frames iteratively. Starting from the first frame, we calculate the similarity with the subsequent frame. If the similarity is quite high, we skip to the next until a frame with significant change is found. We then use this frame as a new reference point and continue to seek subsequent frames with notable differences. The detailed process is provided in~\cref{ssim}. The keyframe-ASR sequence is as follows: $\langle \text{frame}_{1}^{k_1}, \text{frame}_{1}^{k_2}, \text{asr}_1 \rangle,  \text{asr}_2 , \text{asr}_3, \langle \text{frame}_{4}^{k_1}, \text{asr}_4 \rangle, \dots$

\textbf{Keyframe-Level Extraction: Keyframe-to-OCR.} Last but not least, most instructional videos often use bullet-pointed text, formulas, and mathematical symbols to illustrate knowledge points, physical concepts, and calculation processes. These texts, symbols, and mathematical formulas encapsulate substantial knowledge. Therefore, we extract these texts from keyframes as the ASR's supplement. Specifically, we employ two advanced VLMs (InternVL2-40B~\citep{chen2024far}) to perform optical character recognition (OCR) on each keyframe, extracting on-screen text, mathematical symbols, formulas, and other elements. 

\textbf{Keyframe-Level Filtering: Uninformative Keyframe and Redundant OCR.} Despite filtering visual content at multiple levels, some keyframes may still contain low informational scenes, e.g., occlusion. Therefore, we also utilize InternVL2 to score each keyframe after conducting OCR. Additionally, we do not retain all OCR texts, as the OCR from consecutive keyframes is likely to be highly similar or even identical. Therefore, we filter out OCR results that are similar to previous ones.

Lastly, as shown in~\cref{fig2}, through our multi-level extracting and filtering, we curate high-quality video keyframes, OCR text, and ASR transcriptions. These elements represent the useful visual content in videos and the instructor's in-depth explanation of knowledge points. To create the pretraining dataset, we interleave the selected keyframes of a long video with refined ASR and OCR text in chronological order, creating our multimodal textbook: $\{\text{frame}_{1}^{k_1}\! , \text{frame}_{1}^{k_2}\! , \text{ocr}_1\! , \text{asr}_1\! , \text{asr}_2, \text{asr}_3, \!\text{frame}_{4}^{k_1}\! , \text{ocr}_4, \text{asr}_4 ,..\}$

\section{Analysis of Multimodal Textbook}
\subsection{General statistics} 
We utilize GPT-4o to synthesize our knowledge taxonomy with 3915 knowledge points across 6 subjects, which enabled us to automatically collect 159K English instructional videos based on this taxonomy. Following our video-to-textbook pipeline, we filter 53\% low-quality or repetitive videos and retain 75K videos (22,697 class hours) with an average duration of 18 minutes. Then we extract 6.5M keyframes and 0.75B text (ASR+OCR) tokens from these videos. To enhance training efficiency, we concatenate multiple $\langle \text{frame}_{i}^{k_1}, .. ,\text{frame}_{i}^{k_n}, \text{ocr}_i, \text{asr}_i \rangle$ fragment into a single sample, producing a total of 610K interleaved samples. Each sample contains an average of 10.7 keyframes and 1,297 text tokens. The detailed statistics for each subject are shown in Appendix (\cref{Dataset Statistics}). Besides, we randomly select 100 videos and corresponding samples for manual evaluation, with detailed results presented in ~\cref{human evaluation}.

\subsection{Comparison with Existing Datasets} \label{dataset}

\textbf{Image and Text Distribution.} To better demonstrate the advantages of our video-centric dataset, we compare our multimodal textbook with existing datasets (image-text paired datasets and webpage-centric datasets), focusing on the distribution of images and tokens across these datasets. As shown in~\cref{Dataset compare}, we observe that our dataset exceeds previous datasets in terms of the average number of images and text tokens. For instance, our dataset contains an average of 10.7 images per sample, compared to only 5.7 in MMC4 and 4.1 in OBELICS. 

\textbf{Images within a Sample are More Closely Related.} A notable feature of our video-centric design is the inherent association between multiple images within a sample, providing a dynamic illustration of mathematical concepts or physical phenomena. To validate this, we design an in-sample image similarity metric (InSI-SIM). It measures the similarity between all images within a sample, i.e., calculating the average of all pairwise similarity of a sample. For similarity, we consider both semantic (CLIP score) and structural similarity (SSIM score) respectively. The detailed formula is presented in~\cref{InsSI-SIM}.

As shown in~\cref{Dataset compare}, we report $\text{InSI-SIM}$ for 8 image-subset (i.e., the subset containing 4 images) to 8 image-subset ($L$: 4 to 8). For all subsets, our multimodal textbook achieves a significantly higher $\text{InSI-SIM}$ score than other datasets, nearly more than double. For example, our textbook scores 0.686 on average, while OBELICS reaches only 0.345. Besides, we also observed that, as the number of images per sample increases, the $\text{InSI-SIM}$ of our dataset remains stable at around 0.68, whereas other datasets experience a noticeable decline (about $\downarrow$ 10\%). This further validates that our video-centric dataset provides more coherent and contextually related images.

\begin{table*}[t!]
    \centering \scriptsize
    \setlength\tabcolsep{6pt} 
    \begin{tabular}{l |cccc| cccc |cccc |cccc}
     \toprule[1pt]
        \textbf{\#Shot} & 0 & 1 & 2 & 4 & 0 & 1 & 2 & 4 & 0 & 1 & 2 & 4 & 0 & 1 & 2 & 4 \\  \toprule[0.7pt]
        \textbf{Dataset} & \multicolumn{4}{c}{$\text{ScienceQA}^{\text{IMG}}$} &\multicolumn{4}{c}{\text{OKVQA}} & \multicolumn{4}{c}{\text{TextVQA}} &\multicolumn{4}{c}{$\text{TextVQA}^{\text{ocr}}$}    \\ \hline
        \multicolumn{1}{l|}{MMC4} &-&1.6&3.9&11.6 &8.6 &23.6 &21.5 &28.7 &12.1 & 16.2 &16.8 &20.9 &14.5 & 23.9 & 29.9 & 34.7  \\ 
        \multicolumn{1}{l|}{\text{MMC4-Core-ff}} &-&2.1&10.1&10.2 &11.8 &21.2 &25.3 &30.4 &\textbf{13.6} & 18.7 &18.8 &22.1 &\textbf{16.1} &26.6 &28.7 &33.1 \\ 
        \multicolumn{1}{l|}{\text{OBELICS}}&-&2.8&3.0&16.4  &\textbf{13.0} &\textbf{31.7} &35.7 &37.5 &9.2 &26.5 &30.2 &32.2 &11 &30.7 &36.3 &41  \\  
        \rowcolor[HTML]{ECF4FF}
        \multicolumn{1}{l|}{\text{Textbook-6.5M}} &\textbf{26.3} &\textbf{29.4} &\textbf{25.1} & \textbf{37.3} &10.2 &31.2 &\textbf{36.8} &\textbf{39.9} &11.8 &\textbf{26.7} &\textbf{32.1} &\textbf{33.5} & 14.1&\textbf{33.1} &\textbf{36.4} &\textbf{42.8} \\  \hline
        
        \textbf{Dataset}& \multicolumn{4}{c}{\text{MathVista}} & \multicolumn{4}{c}{\text{MathVision}} & \multicolumn{4}{c}{\text{MathVerse}} & \multicolumn{4}{c}{\text{Avg.}}\\ \hline
        
        \multicolumn{1}{l|}{\text{MMC4}} & 20.4 & 30 & 27.9 & 26 & 12.2 & 21.3 &15.5 & 16.1 &8.6 & 19.4 & 21.2 & \textbf{15.9} &10.9 &19.4 &19.5 & 21.9  \\ 
        
        \multicolumn{1}{l|}{\text{MMC4-Core-ff}} &22.5 &33.0 &29.2 &27.8 &13.7 &23.4 &16.3 &17.7 &\textbf{8.6} & 19.9 &\textbf{21.8} &15.2 & 12.3& 20.7& 21.4 & 22.3   \\ 

        \multicolumn{1}{l|}{\text{OBELICS}} &21.6 & 28.5&31.1 &27.6 &13.4 &20.1 &16.8 &14.9 &6.9 &19.4 &20.7 &14 &10.7 & 22.8 & 24.8  &26.2 \\

        \rowcolor[HTML]{ECF4FF}
        \multicolumn{1}{l|}{\text{Textbook-6.5M}} &\textbf{24.3} &\textbf{43.4} &\textbf{33.2} & \textbf{29.2} &\textbf{14.5} &\textbf{25.6} &\textbf{18.2} &\textbf{18.1} &7.7 &\textbf{28.5} &19.8 &14.6 & \textbf{15.5} &\textbf{31.1} & \textbf{28.8} & \textbf{30.8}  \\  

    \bottomrule[1pt]
    \end{tabular}
    \caption{We continued pre-training the base model of LLaVA-1.5-7B using different interleaved datasets. The results are evaluated on 4 common VQA and 3 math-related benchmarks under few-shot settings.}
    \label{main results1}
\end{table*}

\begin{table*}[t!]
    \centering \scriptsize
    \setlength\tabcolsep{5pt} 
    \begin{tabular}{l|ccccc|ccccc}
     \toprule[1pt]
        & \multicolumn{5}{c|}{\emph{Continual Pre-training from Idefics2-8B-base}} &\multicolumn{5}{c}{\emph{Pre-training Idefics2-8B from scratch}} \\ 
        \text{Dataset} & \text{OKVQA} &\text{TextVQA} & \text{MathVista} & \text{MathVison} & \text{MathVerse} & \text{OKVQA} & \text{TextVQA} & \text{MathVista} & \text{MathVison} & \text{MathVerse}\\      \toprule[0.7pt]
        \multicolumn{1}{l|}{MMC4-cf} &54.1 &57.7 &27.8 &14.0 & 17.3 &9.4&25.1&24&13.3&18.3 \\ 
        \multicolumn{1}{l|}{\text{OBELICS}} &54.6 &57.5 &27.6 &14.3 & 17.5&\textbf{10.5}&25.7&24.2 &13.6&17.7 \\  
        \multicolumn{1}{l|}{\text{Textbook-6.5M}} &\textbf{55.1} &\textbf{58.2} &\textbf{29.7} &\textbf{16.2} &\textbf{19.4} & 10.1&\textbf{26.8}&\textbf{26.1}& \textbf{14.4}&\textbf{19.8} \\ 
  
    \bottomrule[1pt]
    \end{tabular}
    \caption{Except for LLaVA, we also pre-train advanced VLMs with multi-image ability (Idefics): continual pretraining from Idefics-8B-base or pre-training from scratch. The evaluations are extended to an 8-shot using randomly selected examples as previous works~\cite{laurenccon2024matters}.}
    \label{main results2}
    \vspace{-0.2cm}
\end{table*}

\section{Experiments}
\subsection{Experimental Settings}
\textbf{Baselines.} We first employ LLaVA-1.5-7B~\citep{liu2024improved} as base models to study the pretraining performance on our dataset and reference datasets (MMC4, OBELICS). For LLaVA-1.5-7B, we apply continual pretraining on its pre-trained model (aligned using 558K paired data). To investigate our dataset more comprehensively, we also pre-train Idefics2-8B model~\cite{laurenccon2024matters} on our dataset, which is an advanced VLM that already supports multi-image and interleaved format input. For the Idefics2-8B, we design two pretraining settings: 1. Training from scratch using the architecture of Idefics2-8B (i.e., Idefics2-8B with randomly initialized projector) and 2. Continual pretraining from the Idefics2-8B-base which is already pre-trained on OBELICS. For a fair comparison, we sample an equivalent number of samples (610K) from MMC4 and OBELICS and apply the same training parameters across all datasets. 

\textbf{Evaluation Methods.} Following OpenFlamingo~\citep{awadalla2023openflamingo} and OmniCorpus~\citep{li2024omnicorpus}, we evaluate the performance of the pre-trained models on two VQA benchmarks (TextVQA~\cite{singh2019towards}, OKVQA~\citep{marino2019ok}), three visual reasoning benchmarks (MathVista, MathVision, MathVision), and ScienceQA-IMG~\cite{lu2022learn}, covering general, OCR, mathematics, and science domains. We compute model accuracy in few-shot settings using either randomly sampled or retrieved examples as previous works~\cite{laurenccon2024matters, yang2022empirical, li2024omnicorpus}.

\subsection{Main Results}
As shown in~\cref{main results1,main results2}, after being pretrained on our Textbook-6.5M, both LLaVA-1.5 and Idefics-8B exhibit significant improvements across seven benchmarks, achieving average gains of +3.2\%, +8.3\%, +4.0\%, and +4.6\% in the 0-shot to 4-shot settings, respectively. Notably, even for cutting-edge VLMs like Idefics2, our multimodal textbook brings an additional improvement of +1.4\%, underscoring rich knowledge content and its high data quality.

\textbf{Our Textbook Brings Improvement on Knowledge-oriented and Reasoning Benchmarks.} In~\cref{main results1}, we observe that our textbook dataset delivers notably greater improvements on knowledge-oriented and reasoning-related benchmarks compared to counterpart datasets. For instance, on ScienceQA, our dataset achieves over a 20\% improvement in both zero-shot and few-shot settings compared to MMC4. Similarly, on math-related benchmarks such as MathVista, which require both mathematical knowledge and visual reasoning capabilities, our dataset demonstrates an average improvement of +5.3\% and +6.4\% compared to OBELICS. This improvement highlights the high quality of our textbook, which distills extensive knowledge from instructional videos into an interleaved textbook.

\textbf{Coherent Video Frame Interleaving with ASR Enhance the In-context learning capabilities.} We observe an interesting phenomenon: even on general-domain benchmarks such as OKVQA and TextVQA, our textbook dataset yields modest improvements in few-shot settings. Specifically, as shown in~\cref{main results1}, in the zero-shot scenario, our textbook lags behind OBELICS by 2.8\%; however, in the 1-shot setting, performance becomes comparable. Notably, in the 2-shot and 4-shot settings, our dataset surpasses OBELICS with improvements of +1.1\% and +2.4\%, respectively. A similar trend can also be observed on the TextVQA. This can be attributed to our video-centric interleaved design, which provides more coherent context and enhances the in-context learning capabilities of VLMs.

\subsection{Analysis}
\textbf{Whether VLMs Can Truly Attend to their Interleaved Context?} To better investigate why our textbook enhances few-shot performance, we design a \textbf{``Cheat Test''}: We replace one of the few-shot examples with the test sample itself and then observe whether the VLMs can notice this \textbf{``cheat shortcut''}. A VLM with strong in-context ability would recognize that its context already contains an identical question and answer, thereby answering the question effortlessly. Therefore, we design a 1-shot and 2-shot ``cheat test''. For the 1-shot ``cheat test'', the prompt contains only one example (\{\textcolor{red}{$I_t$}, \textcolor{red}{$q_t$}, \textcolor{red}{$a_t$}\}) that is identical to the test sample (\{\textcolor{red}{$I_t$}, \textcolor{red}{$q_t$}\}). In 2-shot ``cheat test'', it includes two examples in the prompt: one identical example (\{\textcolor{red}{$I_t$}, \textcolor{red}{$q_t$}, \textcolor{red}{$a_t$}\}) and one random example (\{\textcolor{blue}{$I_t$}, \textcolor{blue}{$q_t$}, \textcolor{blue}{$a_t$}\}). This setup allows us to observe whether the VLMs can allocate sufficient attention to their image-text interleaved context and identify relevant information for question answering.

As shown in ~\cref{Cheat Test}, in both 1-shot and 2-shot scenarios, our dataset significantly outperforms MMC4 and OBELICS by nearly 20\%, particularly on MathVista and MathVision, where we nearly reach 100\% in the 1-shot setting, while MMC4 achieves only 72.6\% and 69.3\%, respectively. Furthermore, from the 1-shot cheat to the 2-shot, the difficulty of cheating increasesas as the context lengthens. As a result, we observe significant performance drops for OBELICS and MMC4 from 1-shot to 2-shot cheating scenarios. However, our textbook dataset only exhibits a smaller drop on most benchmarks and even shows an improvement in OKVQA from 79.2 (1-shot) to 84.3 (2-shot). These results show that VLMs pre-trained with our multimodal textbook can more effectively allocate attention to their interleaved context and capture useful information from longer contexts.

\begin{table}[t!]
    \centering \scriptsize
    \setlength\tabcolsep{3pt} 
    \begin{tabular}{l |cccccc}
     \toprule[1pt]
        \text{Dataset} & \multicolumn{1}{c}{\text{OKVQA}} & \multicolumn{1}{c}{$\text{TextVQA}$} & \multicolumn{1}{c}{\text{Mathvista}} & \multicolumn{1}{l}{\text{Mathvision}} & \multicolumn{1}{c}{\text{Mathverse}}\\ \toprule[0.7pt]
        \rowcolor{gray!13} 
        \multicolumn{6}{l}{\emph{1-shot Cheat: } Example:\{\textcolor{red}{$I_t$},\textcolor{red}{$q_t$},\textcolor{red}{$a_t$}\} + Test-case: \textcolor{red}{$I_t$},\textcolor{red}{$q_t$}}\\

        \multicolumn{1}{l|}{MMC4-cf} &69.0 &41.0 &72.6 &69.3 &55.7  \\ 
        \multicolumn{1}{l|}{\text{OBELICS}} &71.5 &43.8 &67.7 &66.5 &62.8  \\  
        
        \multicolumn{1}{l|}{\text{Ours}} &\textbf{79.2} &\textbf{51.9} &\textbf{94.1} &\textbf{98.4} & \textbf{76.8} \\  \hline
        \rowcolor{gray!13} 
        \multicolumn{6}{l}{\emph{2-shot Cheat: }Example:\{\textcolor{red}{$I_t$},\textcolor{red}{$q_t$},\textcolor{red}{$a_t$}\}, \{\textcolor{blue}{$I_e$},\textcolor{blue}{$q_e$},\textcolor{blue}{$a_e$}\}+Test-case: \textcolor{red}{$I_t$},\textcolor{red}{$q_t$}}  \\  
        \multicolumn{1}{l|}{MMC4-Cf} &53.5 & 39.2&55.7 &51.9 &40.8\\ 
        \multicolumn{1}{l|}{\text{OBELICS}} & 71.3&42.8 &56.7 &39.9&39.5\\  
        
        \multicolumn{1}{l|}{\text{Ours}}& \textbf{84.3}&\textbf{49.4}&\textbf{77.1} &\textbf{70.7} &\textbf{63.1} \\  
  
    \bottomrule[1pt]
    \end{tabular}
    \caption{We design ``\textbf{Cheat Test}'' to observe whether VLMs can attend to their interleaved context. We replace a few-shot example with the test sample itself and observe whether VLM notice this identical $<$image,question,answer$>$ within their prompt. \textcolor{red}{$I_t$}, \textcolor{red}{$q_t$}, \textcolor{red}{$a_t$} denote the test case, \textcolor{blue}{$I_e$}, \textcolor{blue}{$q_e$}, \textcolor{blue}{$a_e$} denote a random selected example.}
    \label{Cheat Test}
    \vspace{-0.3cm}
\end{table}

\begin{figure}[t!]
  \centering
    \includegraphics[width=0.8\linewidth]{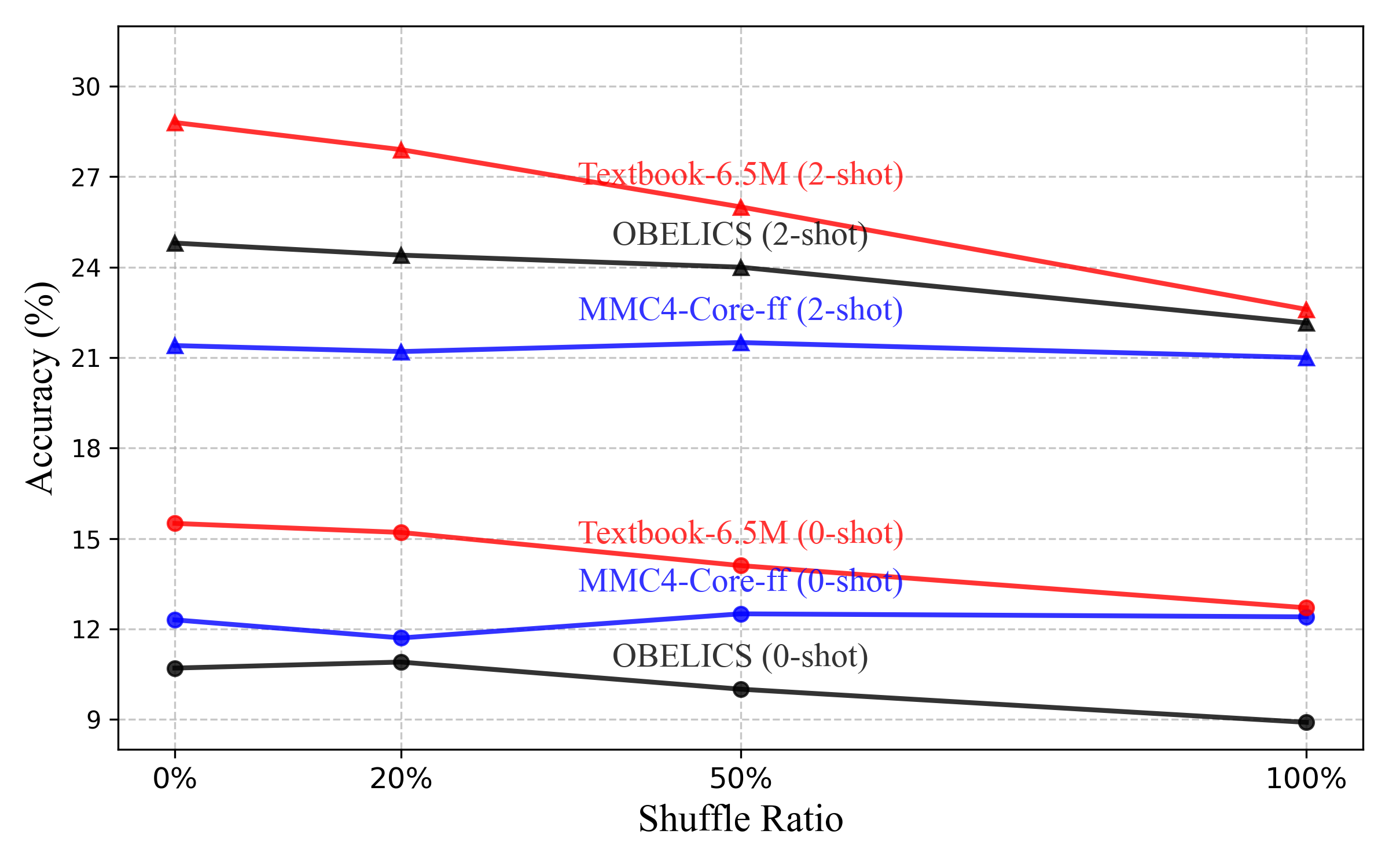}
   \caption{We randomly select 20\%, 50\%, and 100\% samples from datasets and shuffle the image order within each sample. These datasets with shuffled images are also used for pretraining. The Accuracy denotes the average of seven benchmarks.}
   \label{shuffle}
   \vspace{-0.5cm}
\end{figure}

\textbf{The Influence of Disrupting the Image's Order.} 
As previously noted, compared to webpage-centric datasets, the video-centric design offers a more coherent image sequence along with a frame-by-frame text explanatory, presented in an interleaved image-text format. To verify this, we shuffle the image order of interleaved datasets and then also use it for pre-training. For each dataset, we randomly select 20\%, 50\%, and 100\% of the samples and then shuffle the order of images within each sample.

As shown in~\cref{shuffle}, whether shuffled at 20\%, 50\%, or even 100\%, the shuffled MMC4 appears largely unaffected. OBELICS exhibits a moderate decline. In contrast, our multimodal textbook shows a significant performance drop, which becomes increasingly severe as the shuffling ratio increases. These observations confirm our motivation that there is no strong sequential dependency between images in these webpage-centric datasets. However, these coherent images and tightly aligned image-text are beneficial, enabling VLMs to effectively learn complex knowledge and the underlying reasoning logic.

\begin{table}[t!]
    \centering \scriptsize
    \setlength\tabcolsep{4pt} 
    \begin{tabular}{clc|cc}
     \toprule[1pt]
        Pretraining &Continual Pretraining&SFT & \text{OKVQA} & MathVista \\ \toprule[0.7pt]
        \checkmark&  $-$  &\checkmark & 61.1  & 23.2 \\
        \checkmark&  MMC4-Core-ff &\checkmark  & 61.5 \scriptsize$\uparrow$\text{0.4} &24.8 \scriptsize $\uparrow$\text{1.6} \\
        \checkmark&  OBELICS &\checkmark& 61.8 \scriptsize$\uparrow$\text{0.7} &25.6 \scriptsize $\uparrow$\text{2.4} \\
        \checkmark&  Textbook-6.5M &\checkmark  & \textbf{62.2} \scriptsize $\uparrow$\textbf{1.1} & \textbf{28.7} \scriptsize $\uparrow$\textbf{5.5} \\
    \bottomrule[1pt]
    \end{tabular}
    \caption{We also evaluated the zero-shot result after instruction fine-tuning using the 665K data from LLaVA-1.5.}
    \label{SFT results}
\end{table}

\textbf{The Performance after Instruction Turning.} Except for analyzing the pre-training performance, we also report the SFT performance after instruction tuning on LLaVA-665K corpus. All training parameters remain the same for OBELICS, MMC4 and our textbook. As shown in~\cref{SFT results}, on Mathvista, our textbook elevates the performance of the original LLaVA-1.5 from 23.1 to 28.7, achieving an improvement twice (+5.5\%) that of OBELICS (+2.4\%) and three times that of MMC4-Core-ff (+1.6\%). The results of other benchmarks are similar. These results demonstrate that the knowledge learned during pretraining on our multimodal textbook can transfer to instruction fine-tuning stage, leading to positive outcomes for downstream tasks.




\begin{table}[t!]
    \centering \scriptsize
    \setlength\tabcolsep{5pt} 
    \begin{tabular}{lll}
     \toprule[1pt]
        Dataset  & Perplexity $\downarrow$ &1-shot Acc.  \\ \toprule[0.75pt]
         MMC4-Core-ff & 12.56 & 20.7 \\
         OBELICS & 11.27 & 22.8   \\ \hline
         Ours (\emph{ASR Refine, OCR, SSIM})& 13.92 &31.1 \\
         - \emph{w/o ASR Refine} & 16.86 & 26.2 (\scriptsize $\downarrow$4.9)   \\
         - \emph{w/o OCR} & 12.7 & 28.8 (\scriptsize($\downarrow$2.3)\\ \hline
         Keyframe Extraction algorithms & \#Keyframe &1-shot Acc. \\
         - \emph{SSIM$\rightarrow$ Pixel-level extractor}  & 6.5M$\rightarrow$ 18M & 22.1 \scriptsize($\downarrow$9)    \\
         - \emph{SSIM$\rightarrow$ CLIP-based extractor} & 6.5M$\rightarrow$1.7M &24.6 \scriptsize($\downarrow$6.5)   \\
    \bottomrule[1pt]
    \end{tabular}
    \caption{We perform an ablation study on video-to-textbook pipeline, including the impact of ASR refinement, the necessity of incorporating OCR, and the algorithms for extracting keyframes.} \label{ablation study}
    \vspace{-0.5cm}
\end{table}

\subsection{Ablation of Video-to-Textbook's Design}
In ~\cref{pipeline}, we detail the process of our video-to-textbook pipeline, including multi-level extraction and filtering. In this section, we delve into the impact of these designs.

\textbf{Raw ASR Text Impairs the Language Ability.} In our pipeline, we instruct an LLM to refine the transcribed ASR text. As demonstrated in~\cref{ablation study} (\emph{w/o ASR refine}), using raw ASR text results in an average performance drop of 4.9\% across 7 benchmarks.  We calculated the perplexity (PPL) of the raw ASR text and found it significantly higher than other corpora (16.8 Vs. 11.2). This is primarily due to the colloquial characteristics of the video-transcribed ASR, which is often relatively brief, incomplete, and contains a high frequency of meaningless conjunctions. Training directly on such text may impair the model’s language abilities. In contrast, refined ASR has a lower PPL (13.9) and more closely aligns with standard training corpora.

\textbf{Integrating OCR Provides Additional Benefits.} We also analyzed the impact of integrating OCR into our pipeline. The results indicate that OCR provides additional improvements (+2.3\%), particularly in benchmarks such as TextVQA and MathVista. Similar to humans taking notes during lectures, OCR extracts textual knowledge points, formulas, and mathematical symbols from the videos, thereby enhancing the model’s domain-specific expertise. However, we also observed that low-quality OCR can introduce noise and even significantly degrade performance. Therefore, selecting reliable external tools to extract high-quality OCR is crucial.


\textbf{How to Extract Keyframe?} We detect keyframes from video clips using frame-to-frame differences, exploring pixel-level methods (e.g., OpenCV absdiff), structural algorithms (SSIM), and semantic models (CLIP-ViT-L), with results detailed in~\cref{ablation study}. We observed that in these instructional videos, which primarily feature abstract diagrams or geometric images, the pixel-level method often extracts an excessive number of keyframes (18M), resulting in a 9\% drop in training performance. Conversely, the semantic-level model may struggle to distinguish between these geometric images on a semantic level, frequently treating them as similar and consequently missing many critical keyframes (only 1.7M). Therefore, we ultimately adopted SSIM for keyframe extraction, which yielded noticeably better training performance than the other two methods.

\section{Conclusion}
We introduce a multimodal textbook to pre-train VLMs, enabling them to acquire specialized knowledge in a natural and contextual manner. By aggregating online educational videos (e.g., mathematics and physics courses) and transforming them into a frame-ASR interleaved dataset, this textbook provides a coherent and interconnected learning context, complementing traditional image-text alignment methods. Using our pipeline, we curated over 2.5 years of instructional videos (22,000 class hours) into a high-quality dataset with 6.5 million keyframes and 0.75 billion text tokens. Experiments demonstrate its effectiveness, especially in enhancing VLMs' in-context learning capabilities.

{
    \small
    \bibliographystyle{ieeenat_fullname}
    \bibliography{main}

\begin{thebibliography}{60}
\providecommand{\natexlab}[1]{#1}
\providecommand{\url}[1]{\texttt{#1}}
\expandafter\ifx\csname urlstyle\endcsname\relax
  \providecommand{\doi}[1]{doi: #1}\else
  \providecommand{\doi}{doi: \begingroup \urlstyle{rm}\Url}\fi

\bibitem[Abdin et~al.(2024{\natexlab{a}})Abdin, Aneja, Awadalla, Awadallah, Awan, Bach, Bahree, Bakhtiari, Bao, Behl, et~al.]{abdin2024phi3}
Marah Abdin, Jyoti Aneja, Hany Awadalla, Ahmed Awadallah, Ammar~Ahmad Awan, Nguyen Bach, Amit Bahree, Arash Bakhtiari, Jianmin Bao, Harkirat Behl, et~al.
\newblock Phi-3 technical report: A highly capable language model locally on your phone.
\newblock \emph{arXiv preprint arXiv:2404.14219}, 2024{\natexlab{a}}.

\bibitem[Abdin et~al.(2024{\natexlab{b}})Abdin, Aneja, Behl, Bubeck, Eldan, Gunasekar, Harrison, Hewett, Javaheripi, Kauffmann, et~al.]{abdin2024phi}
Marah Abdin, Jyoti Aneja, Harkirat Behl, S{\'e}bastien Bubeck, Ronen Eldan, Suriya Gunasekar, Michael Harrison, Russell~J Hewett, Mojan Javaheripi, Piero Kauffmann, et~al.
\newblock Phi-4 technical report.
\newblock \emph{arXiv preprint arXiv:2412.08905}, 2024{\natexlab{b}}.

\bibitem[Achiam et~al.(2023)Achiam, Adler, Agarwal, Ahmad, Akkaya, Aleman, Almeida, Altenschmidt, Altman, Anadkat, et~al.]{achiam2023gpt}
Josh Achiam, Steven Adler, Sandhini Agarwal, Lama Ahmad, Ilge Akkaya, Florencia~Leoni Aleman, Diogo Almeida, Janko Altenschmidt, Sam Altman, Shyamal Anadkat, et~al.
\newblock Gpt-4 technical report.
\newblock \emph{arXiv preprint arXiv:2303.08774}, 2023.

\bibitem[Alayrac et~al.(2022)Alayrac, Donahue, Luc, Miech, Barr, Hasson, Lenc, Mensch, Millican, Reynolds, et~al.]{alayrac2022flamingo}
Jean-Baptiste Alayrac, Jeff Donahue, Pauline Luc, Antoine Miech, Iain Barr, Yana Hasson, Karel Lenc, Arthur Mensch, Katherine Millican, Malcolm Reynolds, et~al.
\newblock Flamingo: a visual language model for few-shot learning.
\newblock \emph{Advances in neural information processing systems}, 35:\penalty0 23716--23736, 2022.

\bibitem[Awadalla et~al.(2023)Awadalla, Gao, Gardner, Hessel, Hanafy, Zhu, Marathe, Bitton, Gadre, Sagawa, et~al.]{awadalla2023openflamingo}
Anas Awadalla, Irena Gao, Josh Gardner, Jack Hessel, Yusuf Hanafy, Wanrong Zhu, Kalyani Marathe, Yonatan Bitton, Samir Gadre, Shiori Sagawa, et~al.
\newblock Openflamingo: An open-source framework for training large autoregressive vision-language models.
\newblock \emph{arXiv preprint arXiv:2308.01390}, 2023.

\bibitem[Awadalla et~al.(2024)Awadalla, Xue, Lo, Shu, Lee, Guha, Jordan, Shen, Awadalla, Savarese, et~al.]{awadalla2024mint}
Anas Awadalla, Le Xue, Oscar Lo, Manli Shu, Hannah Lee, Etash~Kumar Guha, Matt Jordan, Sheng Shen, Mohamed Awadalla, Silvio Savarese, et~al.
\newblock Mint-1t: Scaling open-source multimodal data by 10x: A multimodal dataset with one trillion tokens.
\newblock \emph{arXiv preprint arXiv:2406.11271}, 2024.

\bibitem[Bai et~al.(2023)Bai, Bai, Yang, Wang, Tan, Wang, Lin, Zhou, and Zhou]{Qwen-VL}
Jinze Bai, Shuai Bai, Shusheng Yang, Shijie Wang, Sinan Tan, Peng Wang, Junyang Lin, Chang Zhou, and Jingren Zhou.
\newblock Qwen-vl: A versatile vision-language model for understanding, localization, text reading, and beyond.
\newblock \emph{arXiv preprint arXiv:2308.12966}, 2023.

\bibitem[Byeon et~al.(2022)Byeon, Park, Kim, Lee, Baek, and Kim]{kakaobrain2022coyo}
Minwoo Byeon, Beomhee Park, Haecheon Kim, Sungjun Lee, Woonhyuk Baek, and Saehoon Kim.
\newblock Coyo-700m: Image-text pair dataset.
\newblock \emph{\url{https://github.com/kakaobrain/coyo-dataset}}, 2022.

\bibitem[Chen et~al.(2024{\natexlab{a}})Chen, Li, Yang, Wen, Yang, Gao, Wu, and Chen]{chen2024comm}
Wei Chen, Lin Li, Yongqi Yang, Bin Wen, Fan Yang, Tingting Gao, Yu Wu, and Long Chen.
\newblock Comm: A coherent interleaved image-text dataset for multimodal understanding and generation.
\newblock \emph{arXiv preprint arXiv:2406.10462}, 2024{\natexlab{a}}.

\bibitem[Chen et~al.(2023)Chen, Wu, Wang, Su, Chen, Xing, Zhong, Zhang, Zhu, Lu, Li, Luo, Lu, Qiao, and Dai]{chen2023internvl}
Zhe Chen, Jiannan Wu, Wenhai Wang, Weijie Su, Guo Chen, Sen Xing, Muyan Zhong, Qinglong Zhang, Xizhou Zhu, Lewei Lu, Bin Li, Ping Luo, Tong Lu, Yu Qiao, and Jifeng Dai.
\newblock Internvl: Scaling up vision foundation models and aligning for generic visual-linguistic tasks.
\newblock \emph{arXiv preprint arXiv:2312.14238}, 2023.

\bibitem[Chen et~al.(2024{\natexlab{b}})Chen, Wang, Tian, Ye, Gao, Cui, Tong, Hu, Luo, Ma, et~al.]{chen2024far}
Zhe Chen, Weiyun Wang, Hao Tian, Shenglong Ye, Zhangwei Gao, Erfei Cui, Wenwen Tong, Kongzhi Hu, Jiapeng Luo, Zheng Ma, et~al.
\newblock How far are we to gpt-4v? closing the gap to commercial multimodal models with open-source suites.
\newblock \emph{arXiv preprint arXiv:2404.16821}, 2024{\natexlab{b}}.

\bibitem[Cheng et~al.(2024)Cheng, Leng, Zhang, Xin, Li, Chen, Zhu, Zhang, Luo, Zhao, et~al.]{cheng2024videollama}
Zesen Cheng, Sicong Leng, Hang Zhang, Yifei Xin, Xin Li, Guanzheng Chen, Yongxin Zhu, Wenqi Zhang, Ziyang Luo, Deli Zhao, et~al.
\newblock Videollama 2: Advancing spatial-temporal modeling and audio understanding in video-llms.
\newblock \emph{arXiv preprint arXiv:2406.07476}, 2024.

\bibitem[Gu et~al.(2024)Gu, Zhang, Zhou, Yu, Xing, Wang, Cao, Jia, Zhang, Wang, et~al.]{gu2024infinity}
Shuhao Gu, Jialing Zhang, Siyuan Zhou, Kevin Yu, Zhaohu Xing, Liangdong Wang, Zhou Cao, Jintao Jia, Zhuoyi Zhang, Yixuan Wang, et~al.
\newblock Infinity-mm: Scaling multimodal performance with large-scale and high-quality instruction data.
\newblock \emph{arXiv preprint arXiv:2410.18558}, 2024.

\bibitem[Gunasekar et~al.(2023)Gunasekar, Zhang, Aneja, Mendes, Del~Giorno, Gopi, Javaheripi, Kauffmann, de~Rosa, Saarikivi, et~al.]{gunasekar2023textbooks}
Suriya Gunasekar, Yi Zhang, Jyoti Aneja, Caio C{\'e}sar~Teodoro Mendes, Allie Del~Giorno, Sivakanth Gopi, Mojan Javaheripi, Piero Kauffmann, Gustavo de Rosa, Olli Saarikivi, et~al.
\newblock Textbooks are all you need.
\newblock \emph{arXiv preprint arXiv:2306.11644}, 2023.

\bibitem[Hu et~al.(2025)Hu, Wu, Pu, Xiao, Zhang, Yue, Li, and Liu]{hu2025video}
Kairui Hu, Penghao Wu, Fanyi Pu, Wang Xiao, Yuanhan Zhang, Xiang Yue, Bo Li, and Ziwei Liu.
\newblock Video-mmmu: Evaluating knowledge acquisition from multi-discipline professional videos.
\newblock \emph{arXiv preprint arXiv:2501.13826}, 2025.

\bibitem[Huang et~al.(2023)Huang, Dong, Wang, Hao, Singhal, Ma, Lv, Cui, Mohammed, Patra, et~al.]{huang2023language}
Shaohan Huang, Li Dong, Wenhui Wang, Yaru Hao, Saksham Singhal, Shuming Ma, Tengchao Lv, Lei Cui, Owais~Khan Mohammed, Barun Patra, et~al.
\newblock Language is not all you need: Aligning perception with language models.
\newblock \emph{Advances in Neural Information Processing Systems}, 36:\penalty0 72096--72109, 2023.

\bibitem[Javaheripi et~al.(2023)Javaheripi, Bubeck, Abdin, Aneja, Bubeck, Mendes, Chen, Del~Giorno, Eldan, Gopi, et~al.]{javaheripi2023phi}
Mojan Javaheripi, S{\'e}bastien Bubeck, Marah Abdin, Jyoti Aneja, Sebastien Bubeck, Caio C{\'e}sar~Teodoro Mendes, Weizhu Chen, Allie Del~Giorno, Ronen Eldan, Sivakanth Gopi, et~al.
\newblock Phi-2: The surprising power of small language models.
\newblock \emph{Microsoft Research Blog}, 1\penalty0 (3):\penalty0 3, 2023.

\bibitem[Jia et~al.(2021)Jia, Yang, Xia, Chen, Parekh, Pham, Le, Sung, Li, and Duerig]{jia2021scaling}
Chao Jia, Yinfei Yang, Ye Xia, Yi-Ting Chen, Zarana Parekh, Hieu Pham, Quoc Le, Yun-Hsuan Sung, Zhen Li, and Tom Duerig.
\newblock Scaling up visual and vision-language representation learning with noisy text supervision.
\newblock In \emph{International conference on machine learning}, pages 4904--4916. PMLR, 2021.

\bibitem[Jiang et~al.(2024)Jiang, He, Zeng, Wei, Ku, Liu, and Chen]{jiang2024mantis}
Dongfu Jiang, Xuan He, Huaye Zeng, Cong Wei, Max Ku, Qian Liu, and Wenhu Chen.
\newblock Mantis: Interleaved multi-image instruction tuning.
\newblock \emph{arXiv preprint arXiv:2405.01483}, 2024.

\bibitem[Lauren{\c{c}}on et~al.(2024{\natexlab{a}})Lauren{\c{c}}on, Marafioti, Sanh, and Tronchon]{laurenccon2024building}
Hugo Lauren{\c{c}}on, Andr{\'e}s Marafioti, Victor Sanh, and L{\'e}o Tronchon.
\newblock Building and better understanding vision-language models: insights and future directions.
\newblock \emph{arXiv preprint arXiv:2408.12637}, 2024{\natexlab{a}}.

\bibitem[Lauren{\c{c}}on et~al.(2024{\natexlab{b}})Lauren{\c{c}}on, Tronchon, Cord, and Sanh]{laurenccon2024matters}
Hugo Lauren{\c{c}}on, L{\'e}o Tronchon, Matthieu Cord, and Victor Sanh.
\newblock What matters when building vision-language models?
\newblock \emph{arXiv preprint arXiv:2405.02246}, 2024{\natexlab{b}}.

\bibitem[Laurençon et~al.(2023)Laurençon, Saulnier, Tronchon, Bekman, Singh, Lozhkov, Wang, Karamcheti, Rush, Kiela, Cord, and Sanh]{laurencon2023obelics}
Hugo Laurençon, Lucile Saulnier, Léo Tronchon, Stas Bekman, Amanpreet Singh, Anton Lozhkov, Thomas Wang, Siddharth Karamcheti, Alexander~M. Rush, Douwe Kiela, Matthieu Cord, and Victor Sanh.
\newblock Obelics: An open web-scale filtered dataset of interleaved image-text documents, 2023.

\bibitem[Li et~al.(2020)Li, Duan, Fang, Gong, and Jiang]{li2020unicoder}
Gen Li, Nan Duan, Yuejian Fang, Ming Gong, and Daxin Jiang.
\newblock Unicoder-vl: A universal encoder for vision and language by cross-modal pre-training.
\newblock In \emph{Proceedings of the AAAI conference on artificial intelligence}, pages 11336--11344, 2020.

\bibitem[Li et~al.(2023{\natexlab{a}})Li, Li, Savarese, and Hoi]{blip2}
Junnan Li, Dongxu Li, Silvio Savarese, and Steven Hoi.
\newblock Blip-2: Bootstrapping language-image pre-training with frozen image encoders and large language models.
\newblock In \emph{International conference on machine learning}, pages 19730--19742. PMLR, 2023{\natexlab{a}}.

\bibitem[Li et~al.(2024)Li, Chen, Wang, Wang, Ye, Jin, Chen, He, Gao, Cui, et~al.]{li2024omnicorpus}
Qingyun Li, Zhe Chen, Weiyun Wang, Wenhai Wang, Shenglong Ye, Zhenjiang Jin, Guanzhou Chen, Yinan He, Zhangwei Gao, Erfei Cui, et~al.
\newblock Omnicorpus: An unified multimodal corpus of 10 billion-level images interleaved with text.
\newblock \emph{arXiv preprint arXiv:2406.08418}, 2024.

\bibitem[Li et~al.(2023{\natexlab{b}})Li, Bubeck, Eldan, Del~Giorno, Gunasekar, and Lee]{li2023textbooks}
Yuanzhi Li, S{\'e}bastien Bubeck, Ronen Eldan, Allie Del~Giorno, Suriya Gunasekar, and Yin~Tat Lee.
\newblock Textbooks are all you need ii: phi-1.5 technical report.
\newblock \emph{arXiv preprint arXiv:2309.05463}, 2023{\natexlab{b}}.

\bibitem[Li et~al.(2023{\natexlab{c}})Li, Zhang, Zhang, Long, Xie, and Zhang]{li2023towards}
Zehan Li, Xin Zhang, Yanzhao Zhang, Dingkun Long, Pengjun Xie, and Meishan Zhang.
\newblock Towards general text embeddings with multi-stage contrastive learning.
\newblock \emph{arXiv preprint arXiv:2308.03281}, 2023{\natexlab{c}}.

\bibitem[Lin et~al.(2023)Lin, Yin, Ping, Lu, Molchanov, Tao, Mao, Kautz, Shoeybi, and Han]{lin2023vila}
Ji Lin, Hongxu Yin, Wei Ping, Yao Lu, Pavlo Molchanov, Andrew Tao, Huizi Mao, Jan Kautz, Mohammad Shoeybi, and Song Han.
\newblock Vila: On pre-training for visual language models, 2023.

\bibitem[Liu et~al.(2023{\natexlab{a}})Liu, Li, Li, and Lee]{liu2023improvedllava}
Haotian Liu, Chunyuan Li, Yuheng Li, and Yong~Jae Lee.
\newblock Improved baselines with visual instruction tuning, 2023{\natexlab{a}}.

\bibitem[Liu et~al.(2023{\natexlab{b}})Liu, Li, Wu, and Lee]{liu2023llava}
Haotian Liu, Chunyuan Li, Qingyang Wu, and Yong~Jae Lee.
\newblock Visual instruction tuning, 2023{\natexlab{b}}.

\bibitem[Liu et~al.(2024{\natexlab{a}})Liu, Li, Li, and Lee]{liu2024improved}
Haotian Liu, Chunyuan Li, Yuheng Li, and Yong~Jae Lee.
\newblock Improved baselines with visual instruction tuning.
\newblock In \emph{Proceedings of the IEEE/CVF Conference on Computer Vision and Pattern Recognition}, pages 26296--26306, 2024{\natexlab{a}}.

\bibitem[Liu et~al.(2024{\natexlab{b}})Liu, Li, Wu, and Lee]{llava}
Haotian Liu, Chunyuan Li, Qingyang Wu, and Yong~Jae Lee.
\newblock Visual instruction tuning.
\newblock \emph{Advances in neural information processing systems}, 36, 2024{\natexlab{b}}.

\bibitem[Lu et~al.(2024)Lu, Liu, Zhang, Wang, Dong, Liu, Sun, Ren, Li, Yang, et~al.]{lu2024deepseek}
Haoyu Lu, Wen Liu, Bo Zhang, Bingxuan Wang, Kai Dong, Bo Liu, Jingxiang Sun, Tongzheng Ren, Zhuoshu Li, Hao Yang, et~al.
\newblock Deepseek-vl: towards real-world vision-language understanding.
\newblock \emph{arXiv preprint arXiv:2403.05525}, 2024.

\bibitem[Lu et~al.(2022)Lu, Mishra, Xia, Qiu, Chang, Zhu, Tafjord, Clark, and Kalyan]{lu2022learn}
Pan Lu, Swaroop Mishra, Tanglin Xia, Liang Qiu, Kai-Wei Chang, Song-Chun Zhu, Oyvind Tafjord, Peter Clark, and Ashwin Kalyan.
\newblock Learn to explain: Multimodal reasoning via thought chains for science question answering.
\newblock \emph{Advances in Neural Information Processing Systems}, 35:\penalty0 2507--2521, 2022.

\bibitem[Marino et~al.(2019)Marino, Rastegari, Farhadi, and Mottaghi]{marino2019ok}
Kenneth Marino, Mohammad Rastegari, Ali Farhadi, and Roozbeh Mottaghi.
\newblock Ok-vqa: A visual question answering benchmark requiring external knowledge.
\newblock In \emph{Proceedings of the IEEE/cvf conference on computer vision and pattern recognition}, pages 3195--3204, 2019.

\bibitem[McKinzie et~al.(2024)McKinzie, Gan, Fauconnier, Dodge, Zhang, Dufter, Shah, Du, Peng, Weers, et~al.]{mckinzie2024mm1}
Brandon McKinzie, Zhe Gan, Jean-Philippe Fauconnier, Sam Dodge, Bowen Zhang, Philipp Dufter, Dhruti Shah, Xianzhi Du, Futang Peng, Floris Weers, et~al.
\newblock Mm1: Methods, analysis \& insights from multimodal llm pre-training.
\newblock \emph{arXiv preprint arXiv:2403.09611}, 2024.

\bibitem[Miech et~al.(2019)Miech, Zhukov, Alayrac, Tapaswi, Laptev, and Sivic]{miech2019howto100m}
Antoine Miech, Dimitri Zhukov, Jean-Baptiste Alayrac, Makarand Tapaswi, Ivan Laptev, and Josef Sivic.
\newblock Howto100m: Learning a text-video embedding by watching hundred million narrated video clips.
\newblock In \emph{Proceedings of the IEEE/CVF international conference on computer vision}, pages 2630--2640, 2019.

\bibitem[Perez et~al.(2021)Perez, Kiela, and Cho]{gpt3}
Ethan Perez, Douwe Kiela, and Kyunghyun Cho.
\newblock True few-shot learning with language models.
\newblock \emph{Advances in neural information processing systems}, 34:\penalty0 11054--11070, 2021.

\bibitem[Radford et~al.(2021)Radford, Kim, Hallacy, Ramesh, Goh, Agarwal, Sastry, Askell, Mishkin, Clark, et~al.]{clip}
Alec Radford, Jong~Wook Kim, Chris Hallacy, Aditya Ramesh, Gabriel Goh, Sandhini Agarwal, Girish Sastry, Amanda Askell, Pamela Mishkin, Jack Clark, et~al.
\newblock Learning transferable visual models from natural language supervision.
\newblock In \emph{International conference on machine learning}, pages 8748--8763. PMLR, 2021.

\bibitem[Sanabria et~al.(2018)Sanabria, Caglayan, Palaskar, Elliott, Barrault, Specia, and Metze]{sanabria2018how2}
Ramon Sanabria, Ozan Caglayan, Shruti Palaskar, Desmond Elliott, Lo{\"\i}c Barrault, Lucia Specia, and Florian Metze.
\newblock How2: a large-scale dataset for multimodal language understanding.
\newblock \emph{arXiv preprint arXiv:1811.00347}, 2018.

\bibitem[Schuhmann et~al.(2021)Schuhmann, Vencu, Beaumont, Kaczmarczyk, Mullis, Katta, Coombes, Jitsev, and Komatsuzaki]{laion400m}
Christoph Schuhmann, Richard Vencu, Romain Beaumont, Robert Kaczmarczyk, Clayton Mullis, Aarush Katta, Theo Coombes, Jenia Jitsev, and Aran Komatsuzaki.
\newblock {LAION-400M:} open dataset of clip-filtered 400 million image-text pairs.
\newblock \emph{CoRR}, abs/2111.02114, 2021.

\bibitem[Schuhmann et~al.(2022)Schuhmann, Beaumont, Vencu, Gordon, Wightman, Cherti, Coombes, Katta, Mullis, Wortsman, et~al.]{schuhmann2022laion}
Christoph Schuhmann, Romain Beaumont, Richard Vencu, Cade Gordon, Ross Wightman, Mehdi Cherti, Theo Coombes, Aarush Katta, Clayton Mullis, Mitchell Wortsman, et~al.
\newblock Laion-5b: An open large-scale dataset for training next generation image-text models.
\newblock \emph{Advances in Neural Information Processing Systems}, 35:\penalty0 25278--25294, 2022.

\bibitem[Singh et~al.(2019)Singh, Natarajan, Shah, Jiang, Chen, Batra, Parikh, and Rohrbach]{singh2019towards}
Amanpreet Singh, Vivek Natarajan, Meet Shah, Yu Jiang, Xinlei Chen, Dhruv Batra, Devi Parikh, and Marcus Rohrbach.
\newblock Towards vqa models that can read.
\newblock In \emph{Proceedings of the IEEE/CVF conference on computer vision and pattern recognition}, pages 8317--8326, 2019.

\bibitem[Sun et~al.(2024)Sun, Cui, Zhang, Zhang, Yu, Wang, Rao, Liu, Huang, and Wang]{sun2024generative}
Quan Sun, Yufeng Cui, Xiaosong Zhang, Fan Zhang, Qiying Yu, Yueze Wang, Yongming Rao, Jingjing Liu, Tiejun Huang, and Xinlong Wang.
\newblock Generative multimodal models are in-context learners.
\newblock In \emph{Proceedings of the IEEE/CVF Conference on Computer Vision and Pattern Recognition}, pages 14398--14409, 2024.

\bibitem[Team et~al.(2024)Team, Georgiev, Lei, Burnell, Bai, Gulati, Tanzer, Vincent, Pan, Wang, et~al.]{team2024gemini}
Gemini Team, Petko Georgiev, Ving~Ian Lei, Ryan Burnell, Libin Bai, Anmol Gulati, Garrett Tanzer, Damien Vincent, Zhufeng Pan, Shibo Wang, et~al.
\newblock Gemini 1.5: Unlocking multimodal understanding across millions of tokens of context.
\newblock \emph{arXiv preprint arXiv:2403.05530}, 2024.

\bibitem[Touvron et~al.(2023)Touvron, Lavril, Izacard, Martinet, Lachaux, Lacroix, Rozi{\`{e}}re, Goyal, Hambro, Azhar, Rodriguez, Joulin, Grave, and Lample]{llama}
Hugo Touvron, Thibaut Lavril, Gautier Izacard, Xavier Martinet, Marie{-}Anne Lachaux, Timoth{\'{e}}e Lacroix, Baptiste Rozi{\`{e}}re, Naman Goyal, Eric Hambro, Faisal Azhar, Aur{\'{e}}lien Rodriguez, Armand Joulin, Edouard Grave, and Guillaume Lample.
\newblock Llama: Open and efficient foundation language models.
\newblock \emph{CoRR}, abs/2302.13971, 2023.

\bibitem[Vasu et~al.(2024)Vasu, Pouransari, Faghri, Vemulapalli, and Tuzel]{datacomp}
Pavan Kumar~Anasosalu Vasu, Hadi Pouransari, Fartash Faghri, Raviteja Vemulapalli, and Oncel Tuzel.
\newblock Mobileclip: Fast image-text models through multi-modal reinforced training.
\newblock In \emph{Proceedings of the IEEE/CVF Conference on Computer Vision and Pattern Recognition}, pages 15963--15974, 2024.

\bibitem[Wang et~al.(2024{\natexlab{a}})Wang, Zhang, Ji, Zhang, Jiang, Wang, Zhu, Wang, Wang, Huang, et~al.]{wang2024pin}
Junjie Wang, Yin Zhang, Yatai Ji, Yuxiang Zhang, Chunyang Jiang, Yubo Wang, Kang Zhu, Zekun Wang, Tiezhen Wang, Wenhao Huang, et~al.
\newblock Pin: A knowledge-intensive dataset for paired and interleaved multimodal documents.
\newblock \emph{arXiv preprint arXiv:2406.13923}, 2024{\natexlab{a}}.

\bibitem[Wang et~al.(2024{\natexlab{b}})Wang, Bai, Tan, Wang, Fan, Bai, Chen, Liu, Wang, Ge, Fan, Dang, Du, Ren, Men, Liu, Zhou, Zhou, and Lin]{Qwen2VL}
Peng Wang, Shuai Bai, Sinan Tan, Shijie Wang, Zhihao Fan, Jinze Bai, Keqin Chen, Xuejing Liu, Jialin Wang, Wenbin Ge, Yang Fan, Kai Dang, Mengfei Du, Xuancheng Ren, Rui Men, Dayiheng Liu, Chang Zhou, Jingren Zhou, and Junyang Lin.
\newblock Qwen2-vl: Enhancing vision-language model's perception of the world at any resolution.
\newblock \emph{arXiv preprint arXiv:2409.12191}, 2024{\natexlab{b}}.

\bibitem[Wang et~al.(2004)Wang, Bovik, Sheikh, and Simoncelli]{wang2004image}
Zhou Wang, Alan~C Bovik, Hamid~R Sheikh, and Eero~P Simoncelli.
\newblock Image quality assessment: from error visibility to structural similarity.
\newblock \emph{IEEE transactions on image processing}, 13\penalty0 (4):\penalty0 600--612, 2004.

\bibitem[Yang et~al.(2024)Yang, Yang, Hui, Zheng, Yu, Zhou, Li, Li, Liu, Huang, et~al.]{yang2024qwen2}
An Yang, Baosong Yang, Binyuan Hui, Bo Zheng, Bowen Yu, Chang Zhou, Chengpeng Li, Chengyuan Li, Dayiheng Liu, Fei Huang, et~al.
\newblock Qwen2 technical report.
\newblock \emph{arXiv preprint arXiv:2407.10671}, 2024.

\bibitem[Yang et~al.(2022)Yang, Gan, Wang, Hu, Lu, Liu, and Wang]{yang2022empirical}
Zhengyuan Yang, Zhe Gan, Jianfeng Wang, Xiaowei Hu, Yumao Lu, Zicheng Liu, and Lijuan Wang.
\newblock An empirical study of gpt-3 for few-shot knowledge-based vqa.
\newblock In \emph{Proceedings of the AAAI conference on artificial intelligence}, pages 3081--3089, 2022.

\bibitem[Yao et~al.(2024)Yao, Yu, Zhang, Wang, Cui, Zhu, Cai, Li, Zhao, He, et~al.]{yao2024minicpm}
Yuan Yao, Tianyu Yu, Ao Zhang, Chongyi Wang, Junbo Cui, Hongji Zhu, Tianchi Cai, Haoyu Li, Weilin Zhao, Zhihui He, et~al.
\newblock Minicpm-v: A gpt-4v level mllm on your phone.
\newblock \emph{arXiv preprint arXiv:2408.01800}, 2024.

\bibitem[Ye et~al.(2024)Ye, Xu, Liu, Hu, Yan, Qian, Zhang, Huang, and Zhou]{ye2024mplug}
Jiabo Ye, Haiyang Xu, Haowei Liu, Anwen Hu, Ming Yan, Qi Qian, Ji Zhang, Fei Huang, and Jingren Zhou.
\newblock mplug-owl3: Towards long image-sequence understanding in multi-modal large language models.
\newblock \emph{arXiv preprint arXiv:2408.04840}, 2024.

\bibitem[Yue et~al.(2024)Yue, Ni, Zhang, Zheng, Liu, Zhang, Stevens, Jiang, Ren, Sun, et~al.]{yue2024mmmu}
Xiang Yue, Yuansheng Ni, Kai Zhang, Tianyu Zheng, Ruoqi Liu, Ge Zhang, Samuel Stevens, Dongfu Jiang, Weiming Ren, Yuxuan Sun, et~al.
\newblock Mmmu: A massive multi-discipline multimodal understanding and reasoning benchmark for expert agi.
\newblock In \emph{Proceedings of the IEEE/CVF Conference on Computer Vision and Pattern Recognition}, pages 9556--9567, 2024.

\bibitem[Zellers et~al.(2022)Zellers, Lu, Lu, Yu, Zhao, Salehi, Kusupati, Hessel, Farhadi, and Choi]{zellers2022merlot}
Rowan Zellers, Jiasen Lu, Ximing Lu, Youngjae Yu, Yanpeng Zhao, Mohammadreza Salehi, Aditya Kusupati, Jack Hessel, Ali Farhadi, and Yejin Choi.
\newblock Merlot reserve: Neural script knowledge through vision and language and sound.
\newblock In \emph{Proceedings of the IEEE/CVF Conference on Computer Vision and Pattern Recognition}, pages 16375--16387, 2022.

\bibitem[Zhang et~al.(2023)Zhang, Li, and Bing]{damonlpsg2023videollama}
Hang Zhang, Xin Li, and Lidong Bing.
\newblock Video-llama: An instruction-tuned audio-visual language model for video understanding.
\newblock \emph{arXiv preprint arXiv:2306.02858}, 2023.

\bibitem[Zhang et~al.(2025)Zhang, Wang, Liu, Huixin, Jiang, Shen, Hou, Zheng, Zhang, Li, et~al.]{zhang2025embodied}
Wenqi Zhang, Mengna Wang, Gangao Liu, Xu Huixin, Yiwei Jiang, Yongliang Shen, Guiyang Hou, Zhe Zheng, Hang Zhang, Xin Li, et~al.
\newblock Embodied-reasoner: Synergizing visual search, reasoning, and action for embodied interactive tasks.
\newblock \emph{arXiv preprint arXiv:2503.21696}, 2025.

\bibitem[Zhu et~al.(2023)Zhu, Chen, Shen, Li, and Elhoseiny]{zhu2023minigpt}
Deyao Zhu, Jun Chen, Xiaoqian Shen, Xiang Li, and Mohamed Elhoseiny.
\newblock Minigpt-4: Enhancing vision-language understanding with advanced large language models.
\newblock \emph{arXiv preprint arXiv:2304.10592}, 2023.

\bibitem[Zhu et~al.(2024)Zhu, Hessel, Awadalla, Gadre, Dodge, Fang, Yu, Schmidt, Wang, and Choi]{zhu2024multimodal}
Wanrong Zhu, Jack Hessel, Anas Awadalla, Samir~Yitzhak Gadre, Jesse Dodge, Alex Fang, Youngjae Yu, Ludwig Schmidt, William~Yang Wang, and Yejin Choi.
\newblock Multimodal c4: An open, billion-scale corpus of images interleaved with text.
\newblock \emph{Advances in Neural Information Processing Systems}, 36, 2024.

\end{thebibliography}
}
\clearpage
\setcounter{page}{1}
\maketitlesupplementary

\section{Detail of Video-to-Textbook Pipeline}\label{detail of pipeline}
\subsection{Implementation Details}
When synthesizing the Knowledge Taxonomy, we utilize \texttt{GPT-4o} to construct the taxonomy. When filtering video at the metadata level, \texttt{GPT-4o} is also employed to review the metadata of the searched videos. During the Video-to-ASR phase, \texttt{Whisper-large-v3} is used to convert audio into text. Then \texttt{Qwen2-72B-Instruct} is applied to refine the raw ASR transcriptions. In the video-level filtering stage, \texttt{DeepSeek-V2} and \texttt{Llama3-70B-Instruct} are used to score each ASR transcription, enabling the filtration of low-quality videos. A video is filtered out if both LLMs determine its ASR does not meet the required standards. After splitting long videos into short clips, we first use \texttt{VideoLlama2-7B} to generate a detailed caption for each video clip. Subsequently, we compute the similarity between the clip's caption and the ASR using \texttt{GTE-Qwen2-7B-Instruct}. Finally, \texttt{InternVL2} is employed to extract and filter OCR from the keyframe. 

\subsection{Human Evaluation} \label{human evaluation}
We randomly sample 100 examples from the multimodal textbook and conduct a manual quality evaluation, focusing on three key aspects: (1) image quality, (2) the connections between different images in a sample and (3) the relevance between texts and images. After manual inspection, we observe that, aside from chemistry, this batch of samples covers five domains: mathematics (31), physics (16), computer science (16), engineering (25), and earth sciences (12). It contains a total of 1,421 images, including 378 slide-style images, 214 lecture-style images, 414 demonstration animations, and 415 natural scenes. Image analysis reveals that only 7\% (72 images) are highly similar, while the remaining images are related to each other but also exhibit clear distinctions. Text-image relevance analysis shows that the attached text (ASR) correctly explains the visual concepts or computational processes presented in the images, with no ambiguity or redundancy.

\begin{algorithm}
\caption{SSIM-Based Key Frame Extraction Algorithm} \label{ssim}
\begin{algorithmic}[1] \small
\Require Frame sequence $\{F_1, F_2, \dots, F_N\}$, similarity threshold $T$
\Ensure Key frame sequence $\{K_1, K_2, \dots\}$

\State $K \gets \{F_1\}$ \Comment{Initialize key frame sequence with the first frame $F_1$}
\State $reference\_frame \gets F_1$ \Comment{Set the reference frame to $F_1$}
\For{$i = 2$ to $N$}
    \State $SSIM \gets \text{CalculateSSIM}(reference\_frame, F_i)$ \Comment{Calculate SSIM between reference frame and frame $F_i$}
    \If{$SSIM < T$}
        \State $K \gets K \cup \{F_i\}$ \Comment{If SSIM is below threshold, add frame $F_i$ as a key frame}
        \State $reference\_frame \gets F_i$ \Comment{Update the reference frame to $F_i$}
    \EndIf
\EndFor
\State \Return $K$ \Comment{Return the sequence of key frames}
\end{algorithmic}
\end{algorithm}

\subsection{Constructing Pretraining Sample} 
After collecting 6.5M keyframes, and 750M refined ASR, and OCR tokens, we can employ various strategies to construct image-text interleaved samples for pre-training. \ding{172} Similar to a webpage-centric dataset, where each webpage is treated as a separate sample, we treat each video as an individual sample. This simple strategy maintains the semantic integrity of a video. However, it also leads to overly long contexts for most samples, as each video contains an average of 86 keyframes, far exceeding the maximum context length supported by most VLMs. \ding{173} As an alternative, we segment a single long video into multiple samples. It can flexibly segment videos based on the maximum context length supported by VLMs. \ding{174} Besides, we directly concatenate multiple video clips i.e., $\langle \text{frame}_{i}^{k_1}, .. ,\text{frame}_{i}^{k_n}, \text{ocr}_i, \text{asr}_i \rangle$, to the maximum context length. This strategy breaks video boundaries, effectively utilizing computational resources. However, mixing multiple video clips within a single sample may adversely affect training performance. Therefore, we insert a specific token: \texttt{End of Video} at the end of each video to mitigate this.

\subsection{Knowledge Taxonomy} \label{Taxonomy}
\begin{figure*}[t!]
  \centering
    \includegraphics[width=1\linewidth]{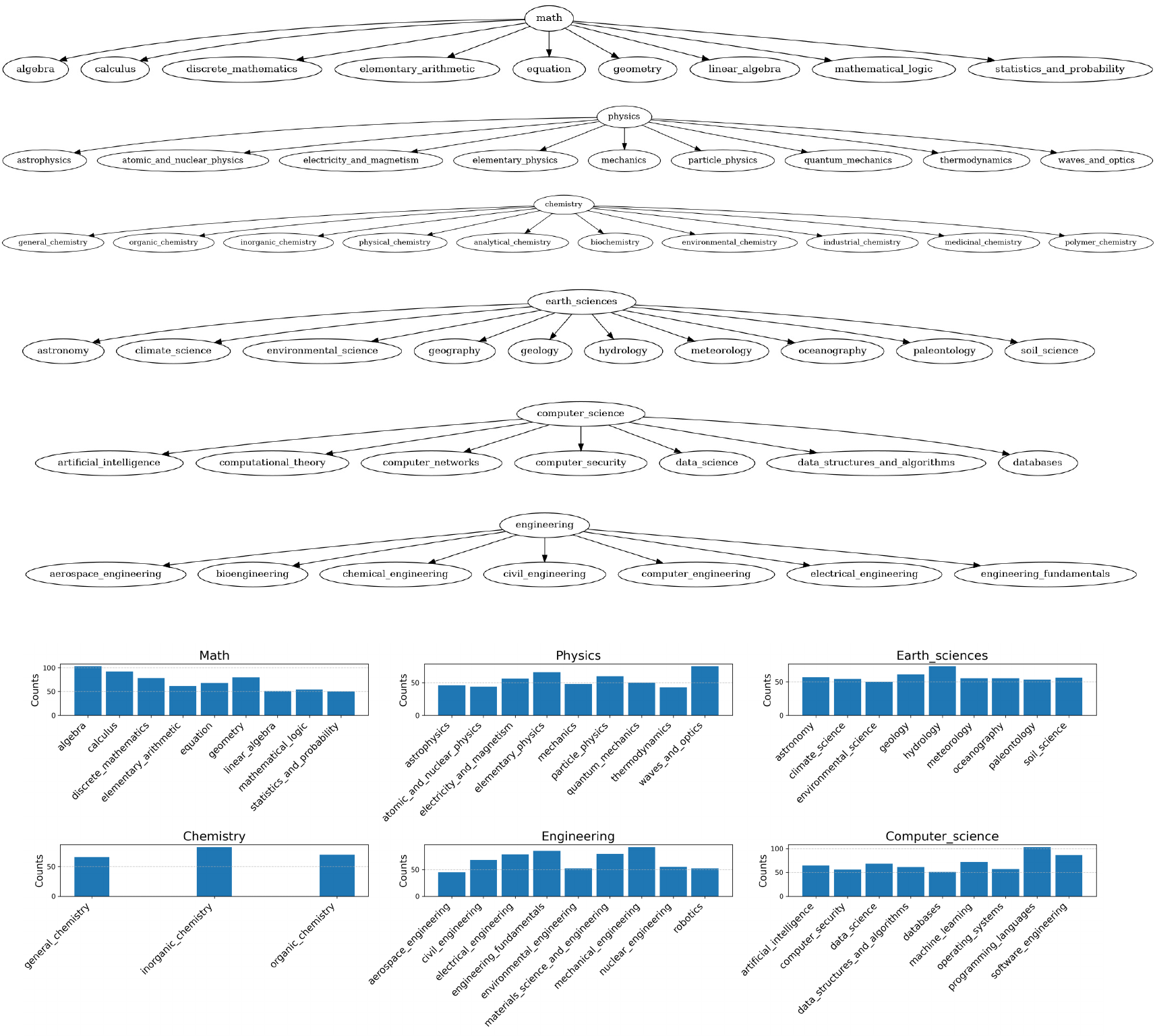}
   \caption{Top: We plot six subjects along with their corresponding sub-courses. Due to space constraints, we selectively visualized only the courses with the highest proportions. Bottom: We count the knowledge points distribution belongs to each subject and its course}
   \label{fig_taxonomy}
\end{figure*}

As stated in the main text, to include richer knowledge in our textbook, we propose a hierarchical knowledge taxonomy comprising four hierarchical layers, namely \emph{Subject $\rightarrow$ Course $\rightarrow$ Sub-course $\rightarrow$ Knowledge Point}. We instruct an LLM to span the knowledge taxonomy across multiple educational stages (from primary school to middle school) and diverse subjects (mathematics, physics, etc.). Lastly, we obtain a knowledge taxonomy comprising 6 subjects (mathematics, physics, chemistry, earth science, engineering, and computer science), 55 courses (Algebra, Solid Geometry,..), and 3915 knowledge points. As illustrated in~\cref{fig_taxonomy}, we plot six subjects along with their corresponding courses. Due to space constraints, we visualized the top 9 courses and their proportion. The number of knowledge points included in each course is approximately the same.

\subsection{Detail of InSI-SIM} \label{InsSI-SIM}
As mentioned in~\cref{dataset}, we design an in-sample image similarity metric (InSI-SIM). It measures the similarity between all images within a sample. Formally, for a subset $D$ containing $M$ samples, each comprising $L$ images, the in-sample image similarity is computed as follows:
\begin{equation}
\text{InSI-SIM}^{L} \!=\! \frac{1}{M} \sum_{k=1}^{M} \frac{1}{\binom{L}{2}} \sum_{i=1}^{L-1} \sum_{j=i+1}^{L} \!\!\Big( \text{CLIP}(\text{Img}_{k,i}, \text{Img}_{k,j}) 
\end{equation}
\begin{equation*}
\quad + \quad  \text{SSIM}(\text{Img}_{k,i}, \text{Img}_{k,j}) \Big) / 2
\end{equation*}
where $\text{CLIP}(\text{Img}_{k,i}, \text{Img}_{k,j})$ and $\text{SSIM}(\text{Img}_{k,i}, \text{Img}_{k,j})$ represent the semantic and structural similarity scores between images $i$ and $j$ in sample $k$, respectively. 

\begin{table*}[t!]
    \centering 
    \small
    \setlength\tabcolsep{5pt} 
    \begin{tabular}{llccccccc}
        \toprule[1pt]
        Subject & \#Video  & Duration (h) & \#Topic & \#Video Clip & \#Keyframe & \#ASR Token & \#OCR Token & \#Sample \\ \hline
        Mathematics      & 21.7k & 4,423& 725 & 809k & 1.67M & 72.5M & 145M  & 123k  \\ 
        Physics          & 11k   & 3,511& 530& 822k  & 0.95M & 36.7M & 73.4M  & 119k  \\ 
        Chemistry        & 4.5k  & 2,643& 410& 234k & 0.49M & 15M & 30M  &  32k \\ 
        Earth Science    & 12k   & 3,670& 520& 640k & 1.03M & 40M & 80M  &  88k \\ 
        Engineering      & 13k   & 4,096& 810& 713k & 1.15M & 43.3M & 86.6M  & 98k \\ 
        Computer Science & 12.8k & 4,354& 820 & 782k & 1.21M & 42.8M & 85.5M  &  150k \\ 
        \textbf{All}     & \textbf{75k} & \textbf{22,697} &\textbf{3,915}  &\textbf{4M} & \textbf{6.58M} & \textbf{258M} & \textbf{500M}  & \textbf{610k} \\
        \bottomrule[1pt]
    \end{tabular}    
    \caption{The statistics of our multimodal textbook. Topic denotes the knowledge points covered by each category of videos, which are sourced from our knowledge taxonomy.}
    \label{Dataset Statistics}
\end{table*}

\lstset{
    framesep = 20pt,
    rulesep = 10pt,
    backgroundcolor = \color[RGB]{245,245,244},
    breaklines = true,
    breakindent = 0pt,
    basicstyle = \ttfamily\small,
    escapeinside = {(*@}{@*)} 
}

\section{Details of Experiments}
\subsection{Detail of Evaluation}
We evaluate the pre-trained VLMs on two VQA benchmarks (TextVQA, OKVQA), a knowledge-centric benchmark (ScienceQA), and three math-related benchmarks (MathVista, MathVerse, MathVision) under few-shot settings. Following the previous works~\cite{li2024omnicorpus}, we use the RICES-based few-shot prompting strategy which retrieves the $k$ most similar samples from the training set based on the testing image feature. It should be noted that since MathVista, MathVerse, and MathVision only contain testing sets, we can not retrieve samples from their respective training sets. Consequently, for MathVista and MathVerse, we retrieve $k$ examples from MathVision, while for MathVision, we retrieve examples from MathVista. When evaluating, we adopt the same prompt as Llava-1.5: 
\begin{lstlisting}
(*@\textbf{System Prompt}@*): A chat between a human and an artificial intelligence assistant. The assistant gives helpful, detailed, and polite answers to the human's questions
(*@\textbf{USER}@*): <image>\n{(*@\color{blue}{example1 query}@*)}\nAnswer the question using a single word or phrase. 
(*@\textbf{ASSISTANT}@*): {(*@\color{blue}{example1 answer}@*)}</s>
(*@\textbf{USER}@*): <image>\n{(*@\color{blue}{example2 query}@*)}\nAnswer the question using a single word or phrase. 
(*@\textbf{ASSISTANT}@*): {(*@\color{blue}{example2 answer}@*)}</s>
....
(*@\textbf{USER}@*): <image>\n{(*@\color{blue}{testing query}@*)}\nAnswer the question using a single word or phrase. 
(*@\textbf{ASSISTANT}@*):
\end{lstlisting}

\subsection{Examples of Multimodal Textbook}
\begin{figure*}[t!]
  \centering
    \includegraphics[width=0.95\linewidth]{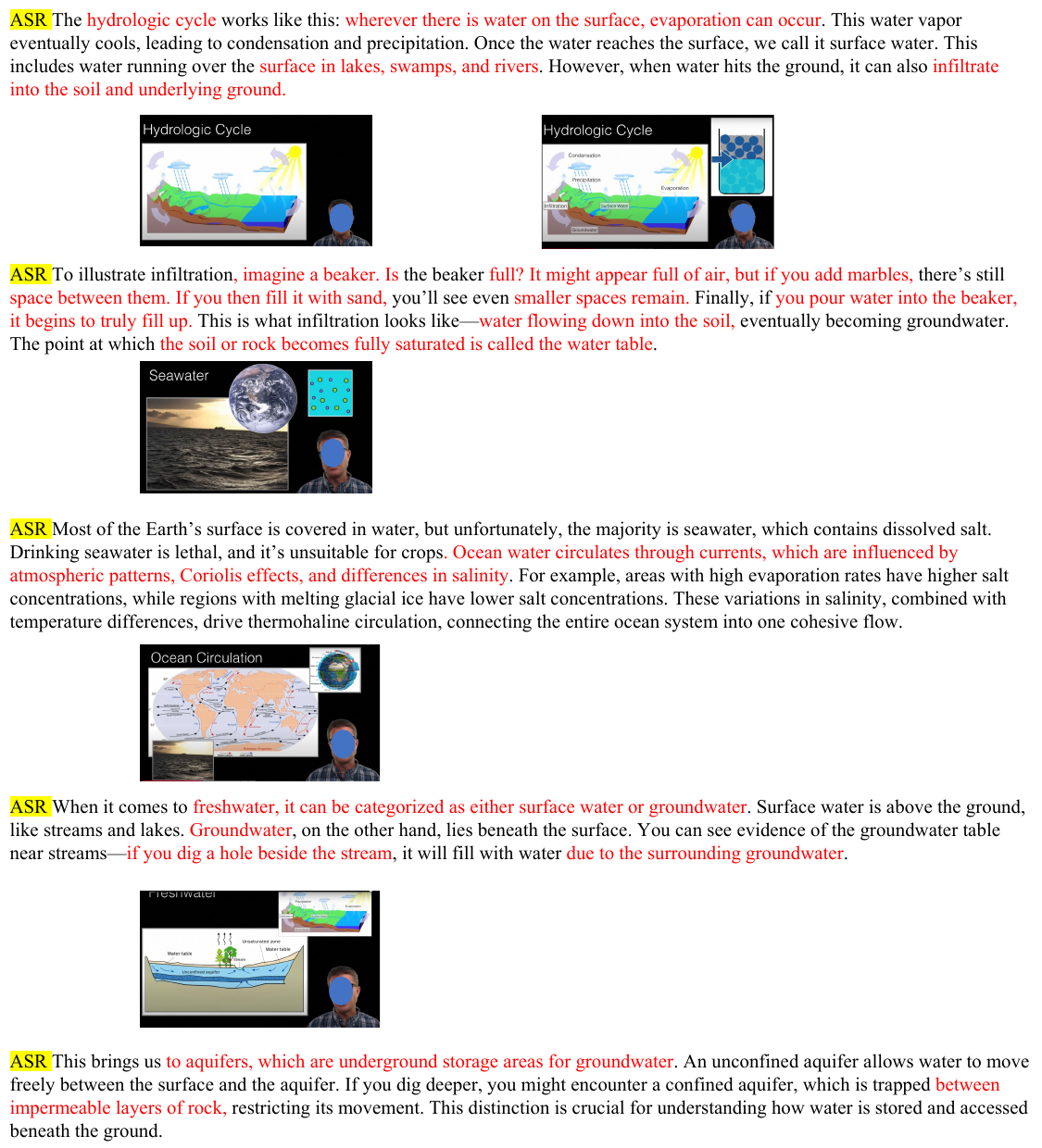}
   \caption{A case presented in our textbook illustrates the water cycle within the domain of earth science.}
   \label{case: earth science}
\end{figure*}

\begin{figure*}[t!]
  \centering
    \includegraphics[width=0.95\linewidth]{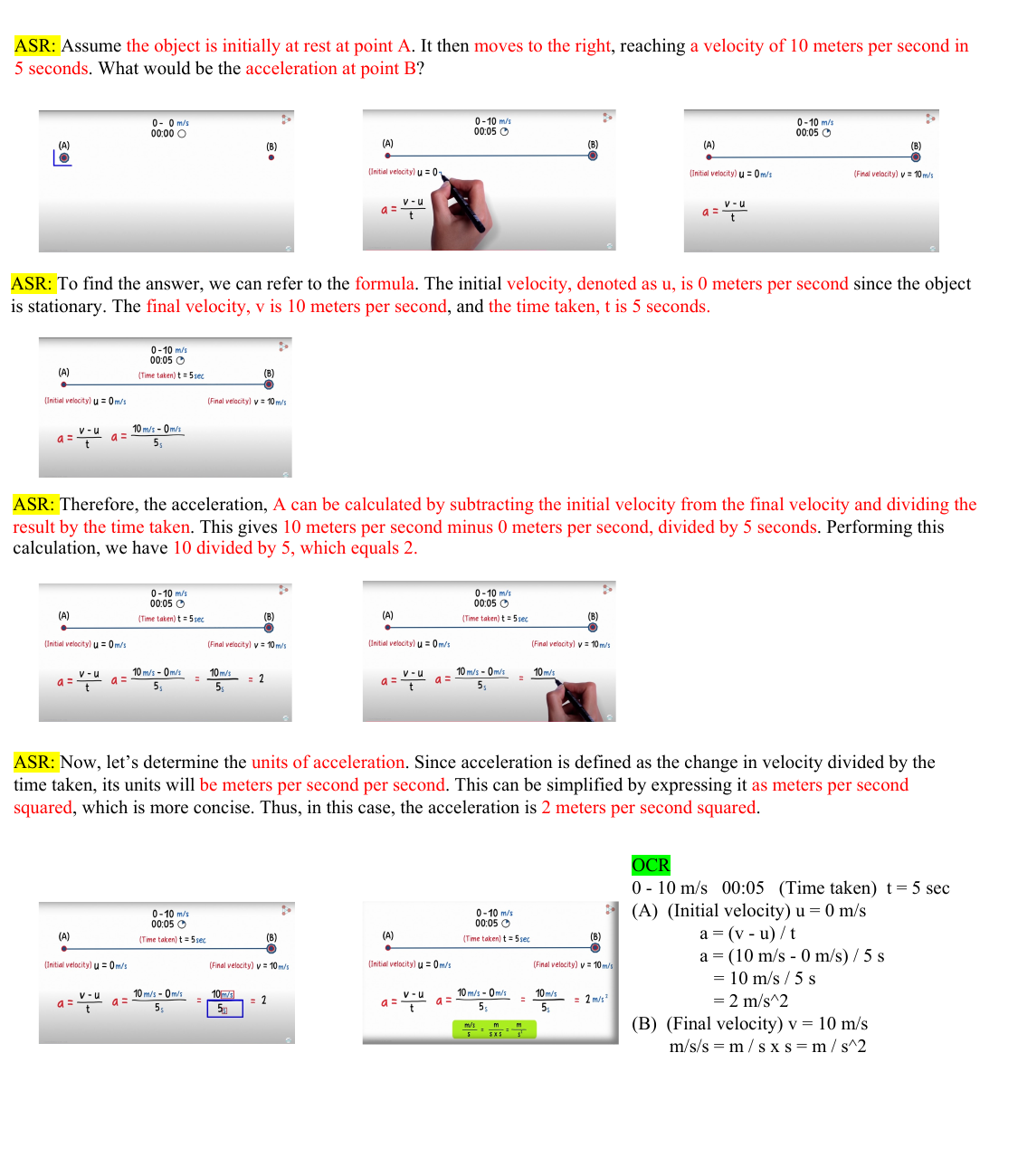}
   \caption{A case presented in our textbook introducing the principles of mechanics within the domain of physics.}
   \label{case: physics1}
\end{figure*}

\begin{figure*}[t!]
  \centering
    \includegraphics[width=0.95\linewidth]{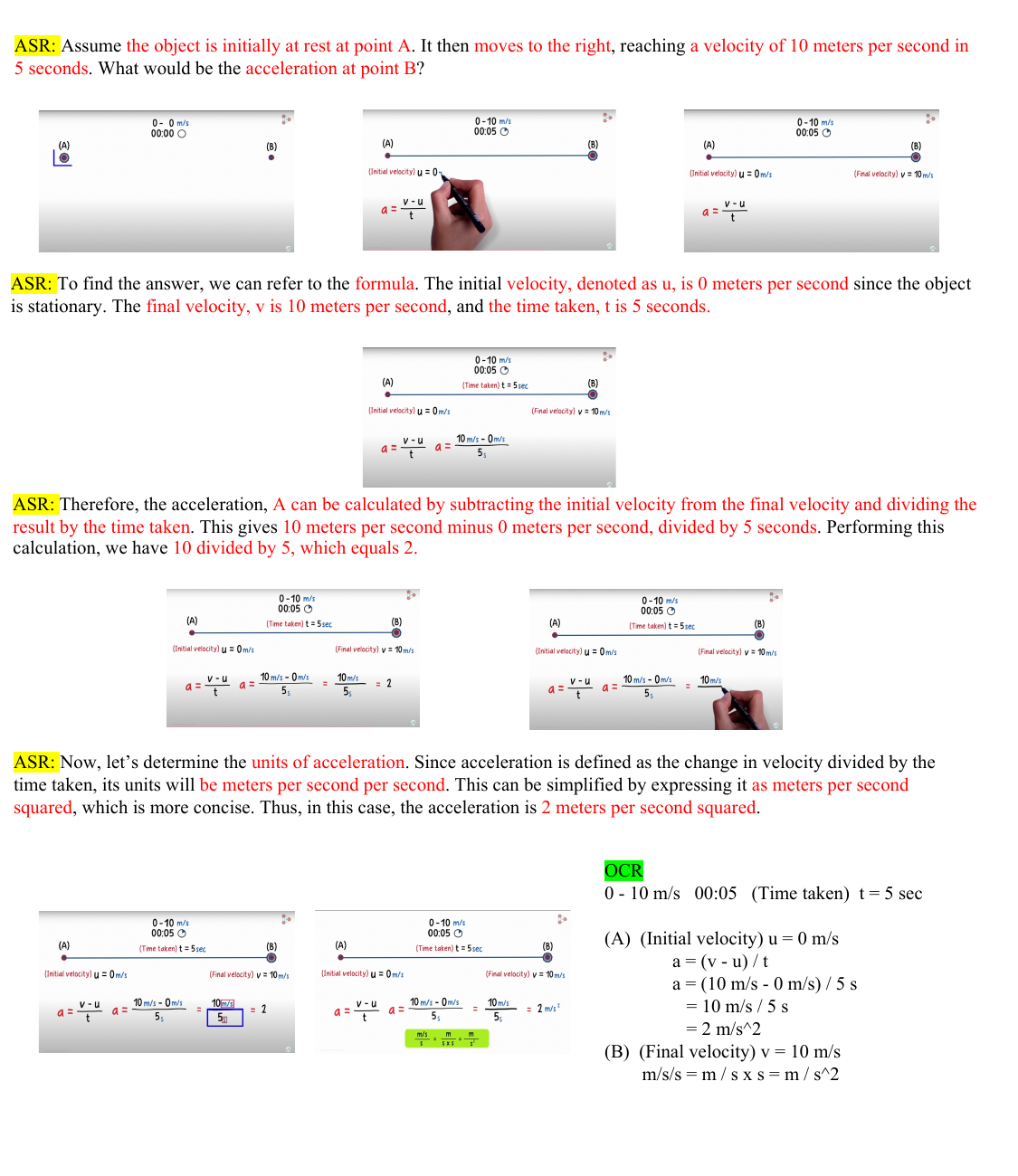}
   \caption{A case presented in our textbook introducing the concepts of velocity and acceleration within the context of physics.}
   \label{case: physics2}
\end{figure*}

\begin{figure*}[t!]
  \centering
    \includegraphics[width=0.95\linewidth]{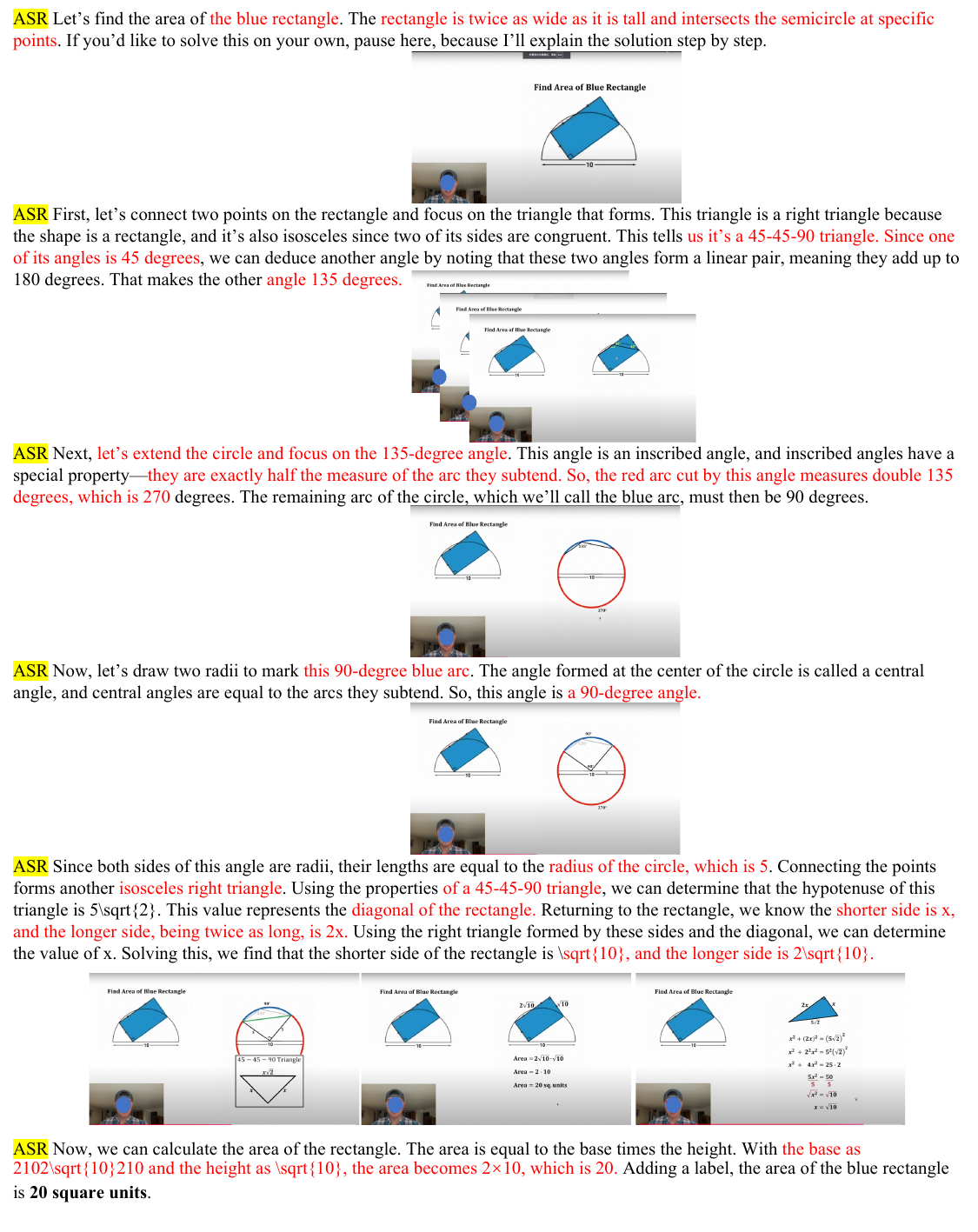}
   \caption{A case presented in our textbook demonstrates how to solve a question about planar geometry in the domain of mathematics.}
   \label{case: math}
\end{figure*}

\begin{figure*}[t!]
  \centering
    \includegraphics[width=0.95\linewidth]{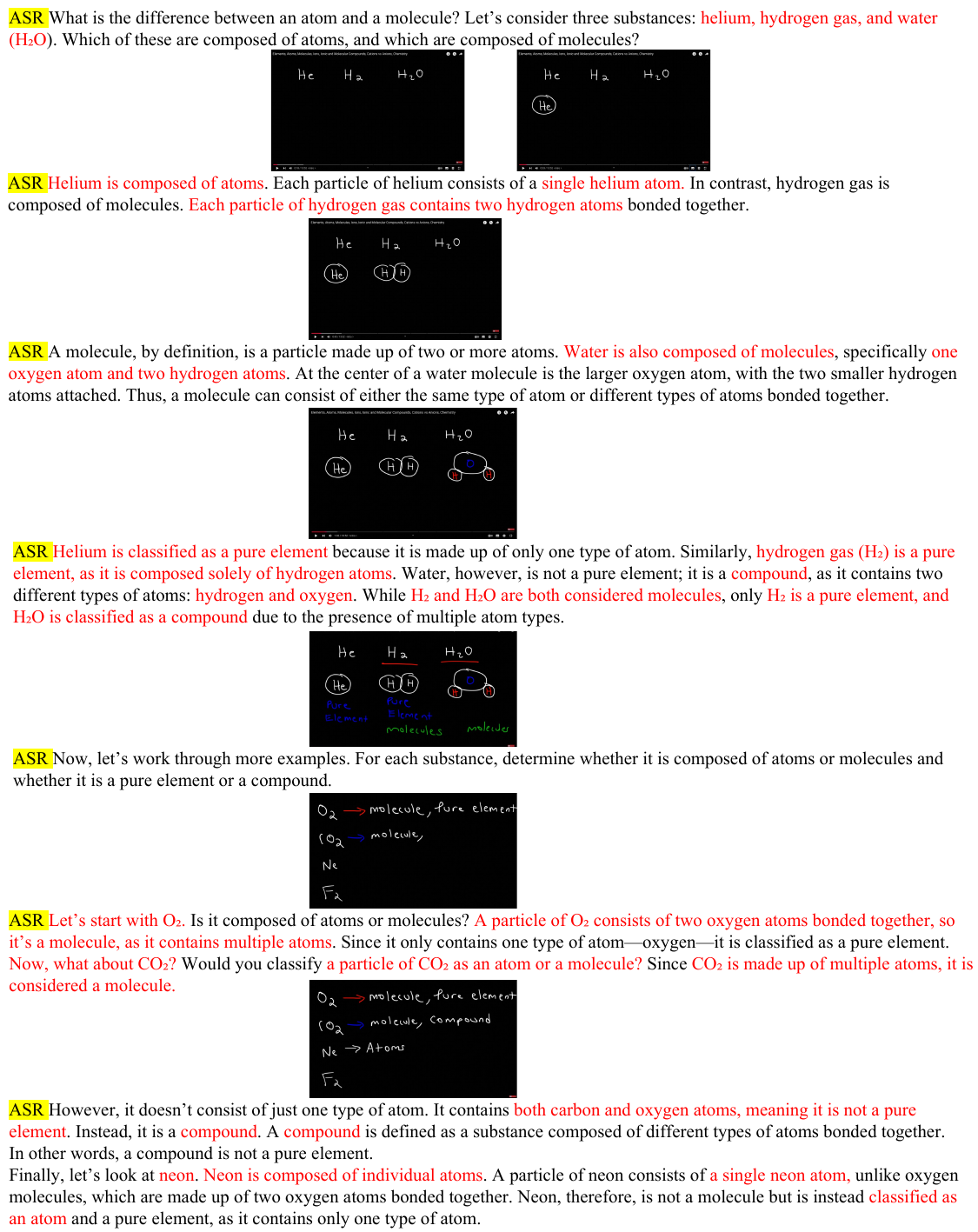}
   \caption{A case presented in our textbook illustrates the concepts of molecules, atoms, and compounds in the domain of chemistry.}
   \label{case: chemistry}
\end{figure*}

\begin{figure*}[t!]
  \centering
    \includegraphics[width=0.95\linewidth]{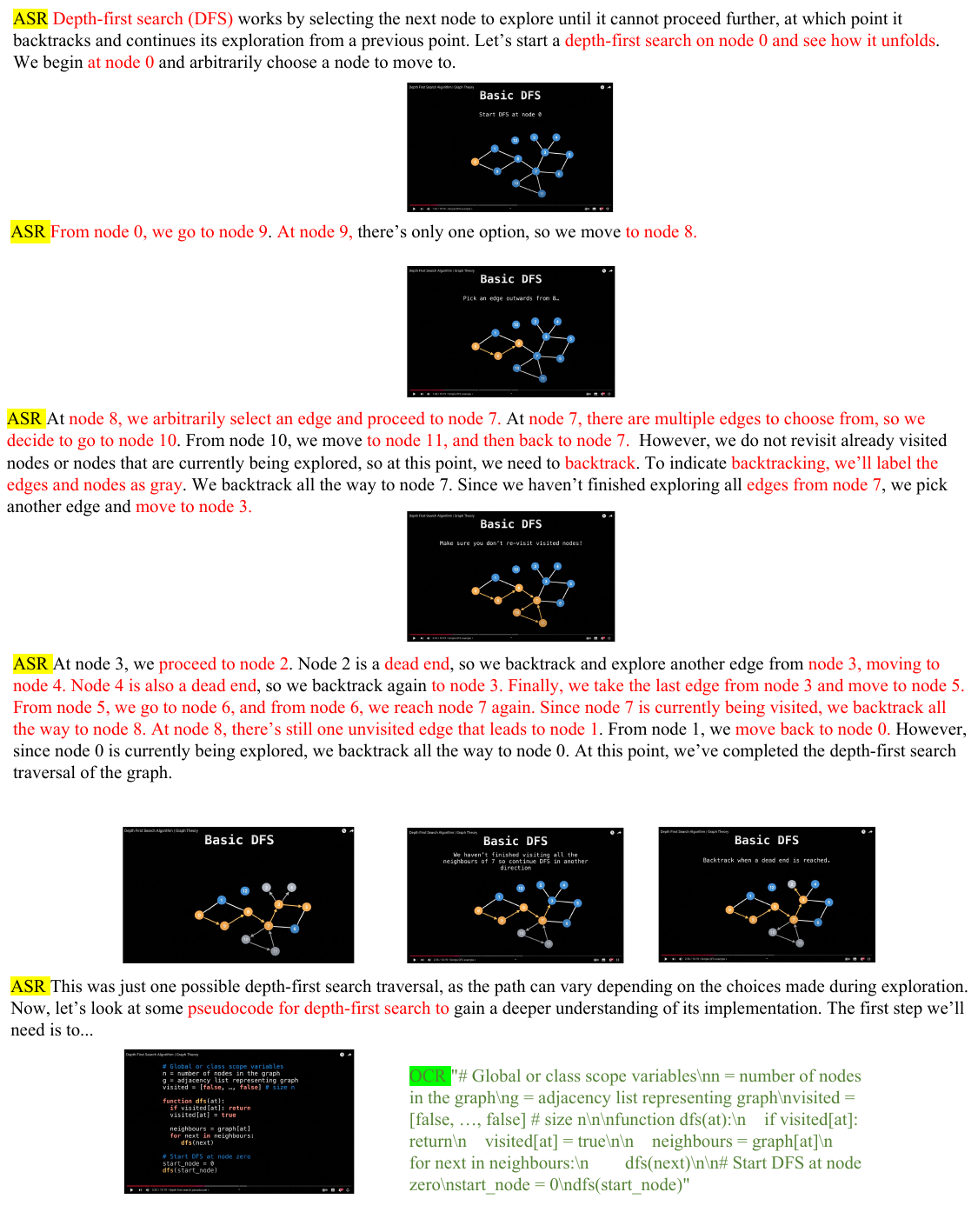}
   \caption{A case presented in our textbook introduces a depth-first search algorithm.}
   \label{case: cs}
\end{figure*}

We provide several detailed examples in~\cref{case: chemistry,case: math,case: earth science,case: cs,case: physics1,case: physics2}. Specifically, \cref{case: earth science} offers a detailed explanation of the Earth's water cycle, presented through slides, photographs, and schematic diagrams. \Cref{case: physics1,case: physics2} provide rich visualizations, including diagrams and texts, to elucidate the concepts of velocity and acceleration in physics. \Cref{case: math} demonstrates the step-by-step, frame-by-frame problem-solving process for a mathematical geometry problem, detailing each critical step with accompanying text and visuals. \Cref{case: chemistry} presents a detailed depiction of chemical concepts such as atoms, molecules, and compounds through a combination of text and illustrations. \Cref{case: cs} introduces the depth-first search algorithm using an animation. 

Except for refined ASR texts, we also provide the OCR texts in our textbook, which can be helpful for math-related scenario. For example, in~\cref{case: physics2}, we utilize OCR to recognize formulas and symbols displayed on the screen, which facilitates better comprehension of physical concepts.

\section{Limitations}
Although we already designed multiple levels of filtering, our textbook may still contain some redundant keyframes, low-quality texts, and so on. We will continue to improve the quality and knowledge density of our textbook. Besides, similar to prior multimodal models, our textbook primarily focuses on multimodal understanding and text generation for interleaved contexts. During training, the loss is not computed for image tokens. However, our textbook can also be used for omni-modal models including both understanding and generation tasks. We leave this for future work.

\section{Ethical discussion}
During the collection and release of our multimodal textbook dataset, We are very concerned about ethical considerations. In addition to following the established corpora (e.g., MMC4~\cite{zhu2024multimodal},
OBELICS~\cite{laurencon2023obelics} and Omnicorpus~\cite{li2024omnicorpus}), we make additional efforts to uphold high ethical standards, such as employing LLMs to filter out inappropriate videos, including those with biases, pornographic content, or personal privacy information, such as identification documents and bank account details. We are open to further refining our strategy while maintaining open-source resources based on
community feedback.

\section{License and Author Statement}
We release the dataset under a CC-BY license and Terms of Use that require disclosure of when the dataset is used for the purpose of training models. This license is not intended to replace the licenses of the source content, and any use of content included in the dataset must comply with the original licenses and applicable rights of its data subjects.

The purpose of this statement is to clarify the responsibilities and liabilities associated with the use of this dataset. While we have made every effort to ensure the accuracy and legality of the data contained within this dataset, we cannot guarantee its absolute completeness or correctness.

Therefore, if any rights, legal or otherwise, are violated through this dataset, including but not limited to copyright infringement, privacy violations, or misuse of sensitive information, we, the authors, assume no liability for such violations.

By utilizing this dataset, you agree that any consequences, legal or otherwise, arising from using this dataset will be the user’s sole responsibility. You acknowledge that you will exercise due diligence and adhere to all applicable laws, regulations, and ethical guidelines when using the dataset.

By accessing, downloading, or using this dataset, you signify your acceptance of this statement and your commitment to abide by the terms and conditions of the CC-BY license.

If you disagree with the terms of this statement or the CC-BY license, you are not authorized to use this dataset.

The dataset will be hosted and maintained on the Hugging Face Hub.

\end{document}